\documentclass[10pt,twocolumn,letterpaper]{article}

\usepackage[pagenumbers]{cvpr}
\definecolor{cvprblue}{rgb}{0.21,0.49,0.74}
\usepackage[pagebackref,breaklinks,colorlinks,allcolors=cvprblue]{hyperref}
\usepackage{footnote}

\title{DyMO: Training-Free Diffusion Model Alignment with Dynamic Multi-Objective Scheduling}

\author{
    Xin Xie,~~Dong Gong\thanks{Corresponding author} \\
    University of New South Wales (UNSW Sydney) \\
  {\tt\small \{xin.xie3, dong.gong\}@unsw.edu.au}
}

\usepackage{algorithm}
\usepackage{algorithmicx}
\usepackage[noend]{algpseudocode}
\usepackage{colortbl}
\usepackage{tcolorbox}

\begin{document}

\twocolumn[{
\renewcommand\twocolumn[1][]{#1}
\maketitle

\centering
\vspace{-0.5cm}
\includegraphics[width=0.9\textwidth]{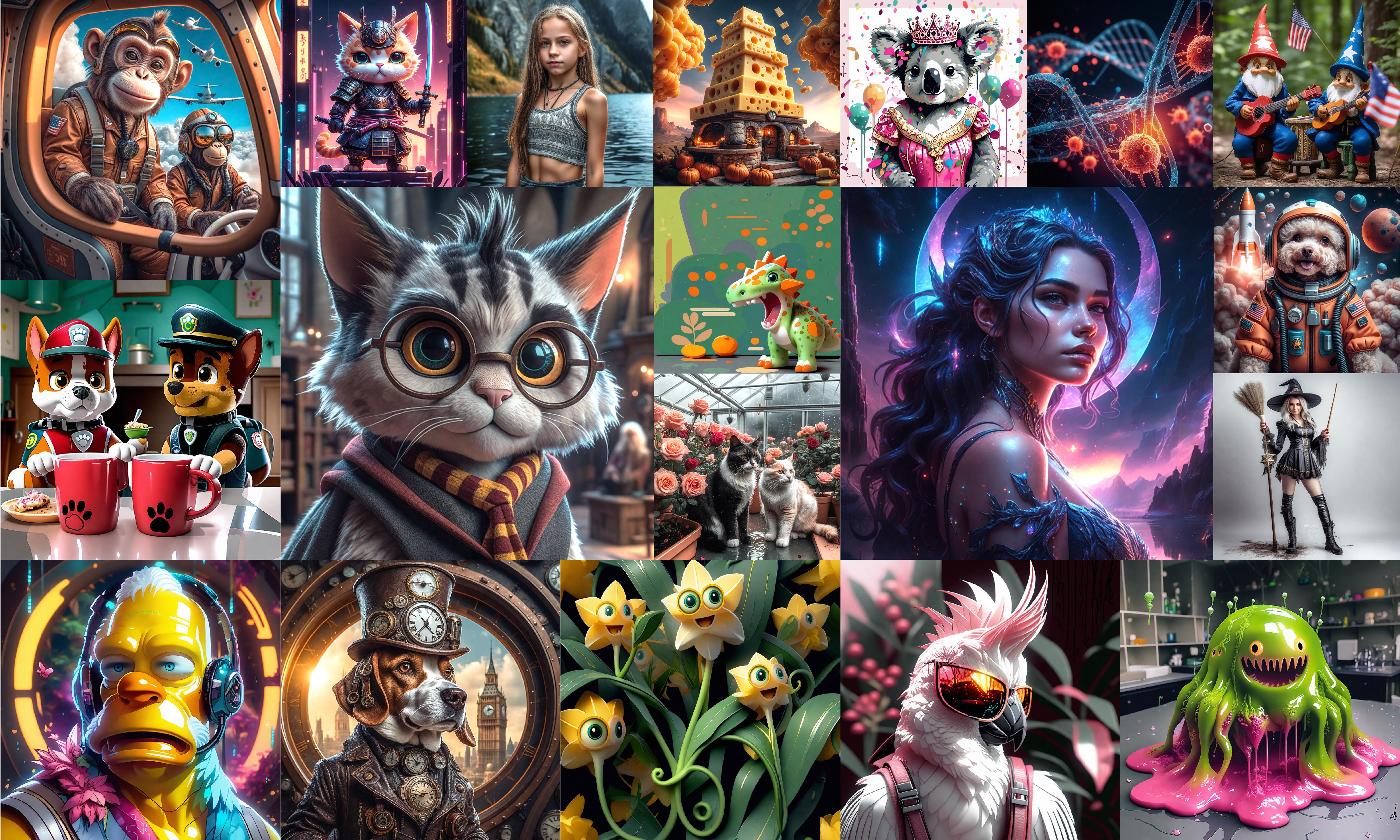}
\vspace{-0.2cm}
\captionsetup{type=figure}
\caption{Sample images generated by our method based on SDXL backbones. Under the guidance of our method, the generated images not only achieve a high alignment with text prompt and human preferences, but also exhibit visually attractive and stunning aesthetics.}
\label{fig:first_image}
\vspace{0.3cm}
}]

\footnotetext[1]{D. Gong is the corresponding author. This project was partially supported by an ARC DECRA Fellowship (DE230101591) to D. Gong.}

\begin{abstract}
Text-to-image diffusion model alignment is critical for improving the alignment between the generated images and human preferences. While training-based methods are constrained by high computational costs and dataset requirements, training-free alignment methods remain underexplored and are often limited by inaccurate guidance. We propose a plug-and-play training-free alignment method, DyMO, for aligning the generated images and human preferences during inference. Apart from text-aware human preference scores, we introduce a semantic alignment objective for enhancing the semantic alignment in the early stages of diffusion, relying on the fact that the attention maps are effective reflections of the semantics in noisy images. We propose dynamic scheduling of multiple objectives and intermediate recurrent steps to reflect the requirements at different steps. Experiments with diverse pre-trained diffusion models and metrics demonstrate the effectiveness and robustness of the proposed method. The project page: \href{https://shelsin.github.io/dymo.github.io/}{dymo.github.io}.
\end{abstract}
\vspace{-0.4cm}
\section{Introduction}
\label{sec:intro}
Text-to-image (T2I) diffusion models \cite{DDPM, SD, SDE, imagen, DALLE3} have demonstrated state-of-the-art effectiveness in image generation, transforming textual prompts into contextually rich visuals. Despite achieving impressive image quality and fidelity, the generative data distribution trained directly from diverse data (\eg, web images) often diverges from user preferences (\eg, aesthetically pleasing images and images aligning the text instructions). To enhance usability, diffusion model \emph{alignment} \cite{Alignment} is investigated to enhance the consistency between the generated images and human user preferences, including the visual preference and intention in the text prompts, through \emph{training-based} \cite{D3PO, DiffusionDPO, DiffusionKTO, SPO, NCPPO, DenseReward, DiffusionRPO} or \emph{training-free} approaches \cite{DOODL, DNO, PromptOpt, ReNO}. 

T2I diffusion model alignment can be achieved through training-based methods \cite{SD, SDXL, SD3, SD3Turbo}, via direct two-stage fine-tuning on customized datasets \cite{HPS, PickAPic} that better represent user preferences with high-quality image-text pairs or preference information. Similarly, relying on text-aware human preference scoring models trained on the preference datasets \cite{HPS, PickAPic}, some methods \cite{DRaFT,AlignProp} directly tune the model to increase the differentiable score.
Rather than straightforward fine-tuning based on static and pre-defined datasets or scores, more advanced and adaptive approaches \cite{DDPO, DPOK} are proposed by incorporating Reinforcement Learning from Human Feedback (RLHF), which allows the diffusion models to learn and refine outputs through reward signals iteratively \cite{PickAPic, Aesthetic, HPS, ImageReward}. To relieve the complexity of learning a reward model, Direct Preference Optimization (DPO) \cite{DPO} that implicitly estimates a reward model is applied to diffusion model alignment \cite{DiffusionDPO, D3PO, SPO, NCPPO}. Despite successful applications, it is challenging and resource-intensive to learn a universal model fulfilling diverse preferences or requirements. 

Training-free methods align generated images with specific objectives by applying differentiable rewards to adjust the generation process of a pre-trained diffusion model during inference \cite{FreeDoM, TFG, DNO, EBAMA}. To enhance text-image alignment, classifier guidance \cite{ClassifierGuidance, SDE} uses the gradient of the pre-trained image classifier. 
The idea is extended to alignment with different types of conditional signals \cite{FreeDoM, TFG, wang2024magic}, and some training-free methods \cite{DOODL, DNO, PromptOpt} also consider aesthetic improvement through optimizing related objectives. 
Unlike training-based methods, most training-free methods are mainly restricted to the alignment of specific conditions (\eg, class). 
To achieve more general applications, in this paper, we consider the \emph{training-free alignment of general preference} in inference. 
The learned text-aware human preference scores \cite{PickAPic, HPS, SPO} measure both text-image alignment and human visual preferences. We can perform training-free alignment with the differentiable scores in score-based diffusion models (SBDMs) \cite{SDE, song2019generative}. Considering that the preference scores perform on clean images by default, specific designs are required to incorporate the guidance to the denoising process in diffusion models. It needs retraining noise-aware or step-dependent score estimator \cite{ClassifierFreeGuidance, GLIDE, MoreControl, UGD} or full-chain backpropagation from the output to intermediate steps \cite{DNO, DOODL, PromptOpt}, which is data or time demanding or time expensive, respectively. 
For memory and time efficiency, we use one-step clean image approximation from noisy image \cite{FreeDoM, UnderstandImprove, TFG}, causing underestimated guidance. Especially, compared to the visual characteristics, the semantic contents (\eg, entities and layout) tend to be determined in early noisy steps. Although the text-aware preference scores can reflect the text-image semantic alignment, it is more challenging to guide the contents due to the blurred predicted image from noisy samples. 

To address these limitations above, we propose a training-free diffusion model alignment approach relying \textbf{Dy}namic scheduling of \textbf{M}ultiple \textbf{O}bjectives (DyMO). We aim to align the generated images with the user preferences in terms of both the intention in the text input and the appealing visual quality. Relying on the pre-trained differentiable text-aware preference scores as an \emph{alignment objective}, we guide the denoising process with the gradient computed for the intermediate noisy images in an SBDM formulation. Since the one-step approximation is used for efficiency and the contents are not sufficiently reflected in early-stage noisy images, the preference model cannot provide effective and accurate guidance. We thus propose a semantic \emph{alignment objective} relying on a discovery that the text-image attention maps are an indirect reflection of the semantic contents (\eg, the entities and layout). The semantic alignment objective/guidance is used to minimize the discrepancies between the attention-map-based contents and the semantic meanings extracted from the text based on a large-language model (LLM). Considering that the two objectives perform with different importance for different stages, we dynamically schedule two objectives to generate detailed content while keeping the layout.  Additionally, we propose a dynamic recurrent strategy to improve the guidance by automatically deciding the number of iterations at different stages. 

The main contributions are summarized as:

\begin{itemize}

    \item We propose a plug-and-play training-free diffusion model alignment method, DyMO, that can effectively align the generated images and human preferences in terms of both user intentions in text and appealing visual quality. 
	
	\item Apart from text-aware human preference scores, we introduce a semantic alignment objective to mitigate the ineffectiveness of the guidance in the early stages of diffusion. We propose dynamic scheduling of multiple objectives and intermediate recurrent steps to reflect the requirements at different steps. 
	
	\item We conduct validation of the proposed DyMO with diverse pre-trained diffusion models, \eg, SD V1.5, SDXL, etc. DyMO outperforms different pre-trained baseline models and other state-of-the-art training-based and training-free methods significantly on different metrics, demonstrating effectiveness and superiority. 

\end{itemize}

\section{Related Work}
\label{sec:related}

\subsection{Diffusion Model Fine-Tuning}
Recently, diffusion models have rapidly advanced in image generation, with various methods emerging to improve quality and align with human preference. One simple yet effective way \cite{AlignProp, ImageReward, DRaFT, DRTune, RAFT} is to use differentiable reward functions as the objective, enabling model optimization through gradient backpropagation. To tackle the limitations of non-differentiable reward functions and unstable training dynamics, some RL-based methods \cite{lee2023aligning, DDPO, DPOK} are proposed for better image-text alignment. Following the success of Direct Preference Optimization (DPO) \cite{DPO} by eliminating the need for explicit reward models, D3PO \cite{D3PO} fine-tuned the model on the preferred and dispreferred image pairs based on human evaluators. Diffusion-DPO \cite{DiffusionDPO} re-formulated DPO for model optimization on the preference dataset\cite{PickAPic}. SPO \cite{SPO} further generalized DPO paradigm at each denoising step to ensure accurate step-aware preference alignment. DenseReward \cite{DenseReward} introduced a temporal discounting factor to prioritize early steps for better preference alignment. NCPPO \cite{NCPPO} emphasized preference optimization in perceptual feature space with more information. Diffusion-RPO \cite{DiffusionRPO} made contrastive weights for image pairs and Diffusion-KTO \cite{DiffusionKTO} defined the alignment objective as human utility maximization. Despite notable progress in diffusion model alignment, discrepancies persist between model performance and human preferences, along with large computational demands.

\subsection{Training-Free Guidance}
Training-free methods can be likened to ``navigating uncharted waters with a compass'', developing various interesting technologies to advance image fidelity. Prafulla \emph{et al.} \cite{ClassifierGuidance} firstly proposed classifier-based guidance to enable conditional image synthesis. Any loss functions can be the classifer, numerous approaches \cite{FreeDoM, TFG, wang2024magic, DOODL, UGD} are proposed to align with different conditional signals by setting related objectives. For example, previous works control the contents by relying on the manipulation on the noise map or attention maps, which are restricted to pre-defined and additionally given layout \cite{FreeDoM, EGGen} or oversimplified the semantics \cite{EBAMA}, limiting them to simple text prompts and restricted usage cases. Additionally, there are some methods \cite{ReNO, InitNO, DNO} designed to optimize injected noise vectors, increasing the differentiable reward score. Similarly, Deckers \emph{et al.} \cite{PromptOpt} conducted prompt embedding manipulation to better align with users’ intentions. Besides, recent works have paid more attention on guidance accuracy and efficiency. Shen \emph{et al.} \cite{UnderstandImprove} utilized random augmentation \cite{RandomAugmentation} to alleviate the adversarial gradient and Polyak step size to accelerate the convergence. Dreamguider \cite{Dreamguider} eliminated compute-heavy backpropagation through the diffusion network by regulating the gradient flow via a time-varying factor. These studies show impressive image generation without any training but are limited to the computational cost and inaccurate guidance, leading to suboptimal outcomes.

\section{Preliminaries}

\begin{figure*}[t]
  \centering
  \includegraphics[width=0.9\linewidth]{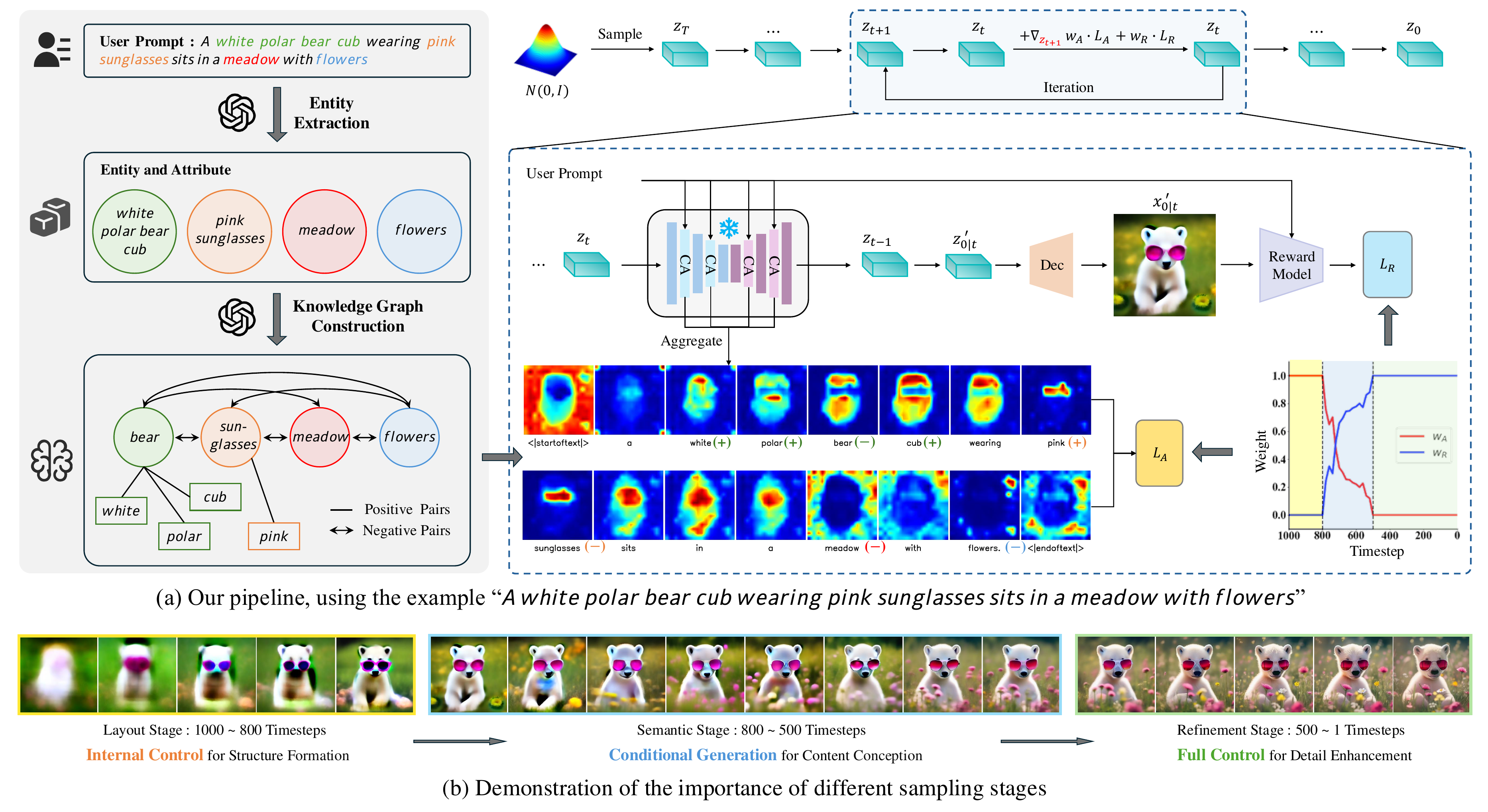}
  \vspace{-0.3cm}
  \caption{The framework of our method. (a) Given a user prompt, we use the LLMs to identify the entities and corresponding attributes for knowledge graph construction. Then we design a semantic alignment objective via cross attention map alignment based on graph, cooperating with a pre-trained preference model to dynamically guide the denoising process for high-quality image generation. (b) The entire denoising process of one-step predicted clean images under the guidance of our method. }
  \vspace{-0.3cm}
\label{fig:framework} 
\end{figure*}

\subsection{Diffusion Models}
Diffusion models \cite{DDPM, SD} are composed of forward and reverse processes. Given a clean data distribution $p(\mathbf{z}_0)$, the forward process incrementally transforms the data $\mathbf{z}_0$ into Gaussian noise over the time interval from $0$ to $T$:
\begin{equation}
	\mathbf{z}_t = \sqrt{\bar{\alpha}_t}\mathbf{z}_0 + \sigma_t \boldsymbol{\epsilon}_{t},
\end{equation}
where $\mathbf{z}_t$ is noisy data, $\bar{\alpha}_t = \prod_{i=1}^{t}(1-\beta_i)$, $\beta_t$ is a increasing linearly noise scheduling function of $t$, $\sigma_t=\sqrt{1 - \bar{\alpha}_t}$ and $\boldsymbol{\epsilon}_{t}\sim\mathcal{N}(0,\mathbf{I})$ represents random noise. Conversely, the reverse process reconstructs the original data by denoising manipulation from time $T$ back to $0$. Diffusion models adopt a neural network $\boldsymbol{\epsilon}_{\boldsymbol{\theta}}$ to predict the noise at each step:
\begin{equation}
\label{eq:ddpm_loss}
\centering
\begin{aligned}
	& \quad \quad \quad \min_{\boldsymbol{\theta}} \mathbb{E}_{\mathbf{z}_t, \boldsymbol{\epsilon}_t, t} \left[ \left\| \boldsymbol{\epsilon}_{\boldsymbol{\theta}}(\mathbf{z}_t, t) - \boldsymbol{\epsilon}_t \right\|_2^2 \right] \\
	& = \min_{\boldsymbol{\theta}} \mathbb{E}_{\mathbf{z}_t, \boldsymbol{\epsilon}_t, t} \left[ \left\| \boldsymbol{\epsilon}_{\boldsymbol{\theta}}(\mathbf{z}_t, t) + \sigma_t \nabla_{\mathbf{z}_t} \log p(\mathbf{z}_t) \right\|_2^2 \right],
\end{aligned}
\end{equation}
where $p(\mathbf{z}_t) \sim \mathcal{N}(\sqrt{\bar{\alpha}_t}\mathbf{z}_0, \sigma_t^2 \mathbf{I})$ is the distribution of $z_t$. The reverse process can be expressed as a score-based ODE in the variance-preserving setting \cite{SDE}, where the score estimator $s(\mathbf{z}_t, t)\approx \nabla_{\mathbf{z}_t} \log p(\mathbf{z}_t)$:
\begin{equation}
	\mathbf{z}_{t-1} = (1 + \frac{1}{2} \beta_t ) \mathbf{z}_t + \beta_t \nabla_{\mathbf{z}_t} \log p(\mathbf{z}_t) + \sqrt{\beta_t} \, \boldsymbol{\epsilon},
\end{equation}
where $\boldsymbol{\epsilon} \sim\mathcal{N}(0,\mathbf{I})$ is randomly sampled Gaussian noise.

\subsection{Diffusion Guidance}
For conditional diffusion models, the target is to generate the data that meets the specified condition $\mathbf{y}$. The reverse process can be reformulated as:
\begin{equation}
\label{eq:cond_sde}
	\mathbf{z}_{t-1} = (1 + \frac{1}{2} \beta_t ) \mathbf{z}_t + \beta_t \nabla_{\mathbf{z}_t} \log p(\mathbf{z}_t|\mathbf{y}) + \sqrt{\beta_t} \, \boldsymbol{\epsilon},	
\end{equation}
where $\nabla_{\mathbf{z}_t} \log p(\mathbf{z}_t|\mathbf{y})$ is the conditional score function. Many recent works, including DPS\cite{DPS}, $\Pi$GDM\cite{IIGDM}, FreeDoM \cite{FreeDoM}, UGD \cite{UGD},  decompose a conditional score function into the unconditional score function and the loss-based term:
\begin{equation}
\label{eq:correction_grad}
\begin{aligned}
	\nabla_{\mathbf{z}_t} \log p(\mathbf{z}_t|\mathbf{y}) &= \nabla_{\mathbf{z}_t} \log p(\mathbf{z}_t) + \nabla_{\mathbf{z}_t} \log p(\mathbf{y}|\mathbf{z}_t) \\
	&= \nabla_{\mathbf{z}_t} \log p(\mathbf{z}_t)- \lambda \nabla_{\mathbf{z}_t} \mathcal{L}_{t}(\mathbf{z}_t, \mathbf{y}),
\end{aligned}
\end{equation}
where $\lambda$ is a scalar coefficient. With \cref{eq:cond_sde} and \cref{eq:correction_grad}, the conditional sampling can be represented as:
\begin{equation}
\label{eq:cond_diffusion}
	\mathbf{z}_{t-1} = \mathbf{m}_t - \eta_t \nabla_{\mathbf{z}_t} \mathcal{L}_{t}(\mathbf{z}_t, \mathbf{y}),
\end{equation}
where $\mathbf{m}_t=(1 + \frac{1}{2} \beta_t ) \mathbf{z}_t + \beta_t \nabla_{\mathbf{z}_t} \log p(\mathbf{z}_t) + \sqrt{\beta_t} \, \boldsymbol{\epsilon}$ and $\eta_t$ is a scaling factor that determines the strength of guidance. In practical applications, the gradient of the last term in \cref{eq:cond_diffusion} is obtained via backpropagation through both the guidance network and the diffusion backbone, accommodating various loss functions.

\vspace{-0.2cm}

\section{Methodology}
\textbf{Overview.}
We propose a training-free framework to align generated contents with contextual semantics and human preferences. We utilize a preference loss $\mathcal{L}_{R}(\mathbf{x}_{0|t}^{\prime}, \mathbf{c}, t)$ to estimate the text-aware human preference score \cite{PickAPic,SPO,ImageReward} of denoised image $\mathbf{x}_{0|t}^{\prime}$ generated by decoding the one-step predicted latent feature $\mathbf{z}_{0|t}^{\prime}$ from $\mathbf{z}_{t}$. We compute the gradient of the reward score to guide the denoising process like \cref{eq:cond_diffusion}. Estimating $\mathcal{L}_R$ and propagating its guidance to early steps is ineffective, where the images are highly noisy or the predicted outputs are highly blurred. To guide image content in the early steps, we leverage text-image attention maps to capture semantic information in the noisy images, as the content is shaped by the attention. We propose a semantic align objective (\cref{subsec:semantic_alignment}), through minimizing the discrepancies between image content semantics (as reflected in attention maps) and text semantics extracted via LLMs. We design the tailored guidance via dynamically scheduling two objectives (\cref{subsec:multi_guidance}), generating detailed content while preserving the layout. Besides, we propose a dynamic time-travel strategy to offer better guidance for diffusion models (\cref{subsec:dynamic_recurrent}). The whole pipeline is demonstrated in \cref{fig:framework} and \cref{algo:code}.

\subsection{Semantic Alignment Guidance}
\label{subsec:semantic_alignment}
\noindent \textbf{Attention map exploration.} During the initial chaotic stage, the basic image structure takes shape quietly. However, the sample $z_t$ exhibits a high level of noise, causing that the approximately predicted image $x_{0|t}^{\prime}$ is extremely blurred. Consequently, training-free guidance is hard to make anything reasonable, providing ineffective semantic supervision. Previous works \cite{EBAMA, Prompt2Prompt, P2P} show attention maps reflect the semantic contents, $M = QK^T/{\sqrt{d}}$, where $d$ is the feature dimension, queries $Q$ and keys $K$ are the linear projection of the intermediate image feature and text embedding of prompt $\mathbf{c}$, respectively. In T2I Diffusion models, each text token $u$ is injected into the image via cross-attention with an attention map $M_u$. We use $M_u$ to supervise early-stage generation by aligning image semantics with the text prompt.

\noindent\textbf{Semantic capture from text prompts.} Upon input of a complex scene prompt $\mathbf{c}$ from the user, we leverage LLMs (\eg, GPT-4 \cite{GPT4} and Llama \cite{Llama}) to analyze the words likely to present in the final image and explicitly extract the entities and their corresponding attributes: $\{E_{i}\}_{i=1}^{n}, \{A_{i}\}_{i=1}^{n}$, where $n$ is the number of the entities identified in the given prompt using LLM and $E_{i}$ represents $i$-th entity, $A_{i}$ denotes the attribute \emph{set} associated with entity node $E_{i}$, encompassing features like color, shape, texture, etc. Each entity possesses a unique set of attributes in varying quantities, which are determined dynamically by LLM-driven heuristics.

\noindent \textbf{Text semantic graph construction.} In order to achieve semantic alignment, we construct a semantic graph $\mathbf{G} = (\mathbf{N}, \mathbf{S})$ to find the internal relationships within the prompts for manipulating attention maps, where $\mathbf{N} = \{E_{i}\}_{i=0}^{n} \cup \{\tilde{A}_{i}\}_{i=0}^{n} $ is the set of nodes in the graph $\mathbf{G}$, where $\tilde{A}_{i}$ denotes the attribute elements in $A_{i}$. For the set of edges $\mathbf{S}$, we propose two kinds of criteria to determine all possible relationships. We hypothesize that each entity exists independently within the image, occupying its own distinct position and space. The relationships between any two entities are \emph{negative}. The relationships between the entities and their corresponding attributes are \emph{positive}, for correct binding between entities and attributes:
\begin{equation}
    \mathbf{S}_{\text{pos}} = \{ (E_i, A_{ij}) \mid \forall i = 1, \cdots, n, \forall j = 1, \cdots, |A_i| \},
\end{equation}
\begin{equation} 
	\mathbf{S}_{\text{neg}} = \{ (E_i, E_m) | i \neq m, \forall i, m = 1, \cdots, n \},
\end{equation}
where $\mathbf{S}_{\text{pos}}$ and $\mathbf{S}_{\text{neg}}$ represent positive and negative pair sets, respectively. $A_{ij}$ denotes the $j$-th attirbute of the $i$-th entity, with a slight abuse of notations. The set of edges is obtained by $\mathbf{S} = \mathbf{S}_{\text{pos}} \cup \mathbf{S}_{\text{neg}}$.

\noindent \textbf{Semantic alignment objective.} Each entity or attribute corresponds to an attention map, denoted by $M_u$. 
We design a semantic alignment loss to align the image contents ($M_u$s' relationships) with the text semantic graph represented by  ($\mathbf{S}_{\text{pos}}$ and $\mathbf{S}_{\text{neg}}$), aiming to achieve better layout composition and precise attribute binding in early steps:
\begin{equation}
\begin{aligned}
	\mathcal{L}_{A} = & -\frac{1}{\left | \mathbf{S}_{\text{pos}} \right | } \sum\nolimits_{(s,l) \in \mathbf{S}_{\text{pos}}}^{} f(M_s, M_l) \\
	& + \frac{1}{\left | \mathbf{S}_{\text{neg}} \right | } \sum\nolimits_{(s,l) \in \mathbf{S}_{\text{neg}}}^{} f(M_s, M_l),
\end{aligned}	
\end{equation}
where $(s,l)$ denotes the edges in $\mathbf{S}$ and $f(\cdot)$ is the cosine similarity function. If more than one tokens are generated from a single object (\eg $u_1, u_2$), this case would be detected, and the attention maps $M_{u_1}, M_{u_2}$ would share the correspondences to the same set of attributes. The proposed semantic alignment loss mitigates the catastrophic object neglect problem by isolating each entity, which prevents the overlap among different objects and helps reduce semantic information loss. Meanwhile, it enhances accurate attribute association by encouraging higher alignment of cross attention maps between objects and related attributes.

\begin{algorithm}[t]
\caption{Our method + Dynamic Time-Travel Straregy}
\begin{algorithmic}[1]
	\Statex \hspace{-1.8em} \textbf{Input:} prompt $\mathbf{c}$, noise predictor $\boldsymbol{\epsilon}_{\boldsymbol{\theta}}(\cdot, t)$, human preference evaluator $\mathcal{L}_{R}(\cdot, \mathbf{c})$, semantic alignment loss function $\mathcal{L}_{A}(\cdot)$, timesteps $T$, decoder $D$, guidance strength $\eta_t$ and pre-defined parameters $\beta_t$, $\bar{\alpha}_t$, $h_t$, $k$.
	\State $\mathbf{z}_{T}\sim\mathcal{N}(0,\mathbf{I})$
	\For{$t = T, ..., 1$}
		\State $\boldsymbol{\epsilon}_1\sim\mathcal{N}(0,\mathbf{I})$ if $t>1$, else $\boldsymbol{\epsilon}_1=0$.
		\State $\tilde{\boldsymbol{\epsilon}}_{t}, M = \boldsymbol{\epsilon}_{\boldsymbol{\theta}}(\mathbf{z}_{t}, t)$
		\State $\mathbf{z}_{t-1} = (1+\frac{1}{2}\beta_t)\mathbf{z}_{t}+\beta_t \tilde{\boldsymbol{\epsilon}}_{t}+\sqrt{\beta_t}\boldsymbol{\epsilon}_1$ 
		\State $\mathbf{z}_{0|t}^{\prime}=\frac{1}{\sqrt{\bar{\alpha}_t}}(\mathbf{z}_t+(1-\bar{\alpha}_t)\tilde{\boldsymbol{\epsilon}}_{t})$
		\State $\mathbf{x}_{0|t}^{\prime}=D(\mathbf{z}_{0|t}^{\prime})$

		\State  $\begin{cases}
          w_A=1, w_R=0  & \text{\textbf{if} \textcolor[HTML]{FF7F50}{$t \ge 800$}} \\ 
          \begin{aligned}
          &w = 1-e^{-k (\frac{\left\| \mathbf{z}_{0|t}^{\prime} - \mathbf{z}_{0|t+1}^{\prime} \right\|}{\left\| \mathbf{z}_{0|t+1}^{\prime} \right\|})} \\
          &w_A= w, w_R = 1-w 
          \end{aligned} & \text{\textbf{if} \textcolor[HTML]{1E90FF}{$ 800 > t \ge 500$}} \\
          w_A=0, w_R=1  & \text{\textbf{if} \textcolor[HTML]{008000}{$ 500 > t \ge 1$}} \\
        \end{cases}$
        
        \State $\mathcal{L} = w_A \cdot \mathcal{L}_{A}(M) + w_R \cdot \mathcal{L}_{R}(\mathbf{x}_{0|t}^{\prime}, \mathbf{c}, t)$

		\State $\boldsymbol{g}_t = \nabla_{\mathbf{z}_t}\mathcal{L}$
		\State $\mathbf{z}_{t-1} = \mathbf{z}_{t-1} - \eta_t \cdot \frac{\left\| \tilde{\boldsymbol{\epsilon}}_{t} \right\|}{\left\| \boldsymbol{g}_t \right\|_2^2} \cdot \boldsymbol{g}_t$
		\State $r_t = h_t \cdot \left\| \boldsymbol{g}_t \right\|$  \Comment{Compute once at each timestep}
		\For{$i = r_t, ..., 1$}  \Comment{Iterate $r_t$ times}
			\State $\boldsymbol{\epsilon}_2\sim\mathcal{N}(0,\mathbf{I})$
            \State $\mathbf{z}_{t} = \sqrt{1 - \beta_t} \mathbf{z}_{t-1} + \sqrt{\beta_t}\boldsymbol{\epsilon}_2$ 
            \State $\text{Repeat from step3 to step16}$  
        \EndFor
	\EndFor
	\State \textbf{return} $x_{0}$
\end{algorithmic}
\label{algo:code}	
\end{algorithm}

\subsection{Multi-Objective Dynamic Scheduling}
\label{subsec:multi_guidance}
In this section, we dynamically guide the entire denoising process and design tailored guidance based on the characteristics of diffusion models to gradually synthesize the high-quality image from $T$ to $0$.

\noindent \textbf{Preference alignment.} After the overall layout of the generated image is roughly established, we incorporate human feedback to guide conditional image generation, aligning the output with human preference in intermediate generation steps. Specifically, we adopt a pre-trained preference model \cite{PickAPic,SPO} as the reward function $\mathcal{L}_{R}(\mathbf{x}_{0|t}^{\prime}, \mathbf{c}, t)=\text{exp}(\tau \cdot f_{\text{V}}(\mathbf{x}_{0|t}^{\prime}, t)^T f_{\text{T}}(\mathbf{c}))$ to update the $\mathbf{z}_{t-1}$ as \cref{eq:cond_diffusion}, where $\tau$ is the temperature and $f_\text{V/T}$ represents vision/text encoder of a CLIP-style model trained on preference data. However, directly applying the preference loss can unintentionally alter the layout because the predicted image $\mathbf{x}_{0|t}^{\prime}$ remains blurred, which leads to inaccurate feedback for the preference model. To address this issue, we propose a solution by adaptively combining the semantic alignment loss and preference alignment loss:
\begin{equation}
\label{eq:dy_w}
\centering
\begin{aligned}
        & \quad w = 1-e^{-k ({\left\| \mathbf{z}_{0|t}^{\prime} - \mathbf{z}_{0|t+1}^{\prime} \right\|}/{\left\| \mathbf{z}_{0|t+1}^{\prime} \right\|})}, \\ 
	& \mathcal{L} = w_{A} \cdot \mathcal{L}_{A}(M) + w_{R} \cdot \mathcal{L}_{R}(\mathbf{x}_{0|t}^{\prime}, \mathbf{c}, t),
\end{aligned}
\end{equation} 
where $w_{A} = w$ and $w_{R} = 1-w$ are dynamical weights and $\mathbf{z}_{0|t}^{\prime}$ denotes the predicted latent features of $\mathbf{x}_{0|t}^{\prime}$. As the image content becomes increasingly stable, the value of weight $w$ decreases and the effect of human feedback increases. The adaptive mechanism for balancing these two losses maintains the crucial image structure while improving the final image quality.

\noindent \textbf{Detail refinement.} In the late stage, as changes in the generated results become minimal, we apply the preference loss $\mathcal{L}_{R}(\mathbf{x}_{0|t}^{\prime}, \mathbf{c}, t)$ to focus on enhancing fine details, such as  texture refinement and subtle feature improvements. These nuanced adjustments boost realism and fidelity, ensuring that the generated output achieves a higher level of texture details and visual aesthetics.

\subsection{Improving Training-free Guidance}
\label{subsec:dynamic_recurrent}
\textbf{Polyak Step Size.} We dynamically schedule the step size of the update of the latent variable via Polyak step size \cite{UnderstandImprove} $\mathbf{z}_{t-1} = \mathbf{z}_{t-1} - \eta_t \cdot \frac{\left\| \tilde{\boldsymbol{\epsilon}}_{t} \right\|}{\left\| \boldsymbol{g}_t \right\|_2^2} \cdot \boldsymbol{g}_t$, where $\eta_t$ is the guidance strength, $\tilde{\boldsymbol{\epsilon}}_{t}$ is the output of $\boldsymbol{\epsilon}_{\boldsymbol{\theta}}(\mathbf{z}_{t}, t)$ and $\boldsymbol{g}_t = \nabla_{\mathbf{z}_t}\mathcal{L}$ represents the gradient of loss function in each stage.

\noindent \textbf{Dynamic Time-Travel Straregy.} After rethinking the efficient time-travel strategy from FreeDom \cite{FreeDoM}, we discover the fixed iteration steps hinder convergence efficiency since the abilities of each denosing step in diffusion models are different. We propose dynamic recurrent scheme to dynamically schedule the number of iterations at each denosing step in diffusion models are different. We compute the iteration count $r_t = h_t \cdot \left\| \boldsymbol{g}_t \right\|$ only once at the start of each step where $h_t$ is a pre-defined parameter. When $\left\| \boldsymbol{g}_t \right\|$ is larger, more iterations are required to promote convergence.

\begin{table}[t]
\centering
\small
\renewcommand{\arraystretch}{1.0}
\caption{Comparison of AI feedback on SD V1.5-based methods.}
\vspace{-0.2cm}
\label{tab:SD15_Compare_Metric}
\setlength{\tabcolsep}{1.1mm}{
\begin{tabular}{c|cccc}
\hline
Methods       & PickScore & HPSv2  & ImageReward & Aesthetics \\ \hline
SD V1.5      & 20.73     & 0.2341 & 0.1697      & 5.337      \\ \hline
DNO          & 20.05     & 0.2591 & -0.3212     & 5.597      \\
PromptOpt    & 20.26     & 0.2490 & -0.3366     & 5.465      \\
FreeDom      & 21.96     & 0.2605 & 0.3963      & 5.515      \\ \hline
AlignProp    & 20.56     & 0.2627 & 0.1128      & 5.456      \\
Diffusion-DPO & 20.97     & 0.2656 & 0.2989      & 5.594      \\
Diffusion-KTO & 21.15     & 0.2719 & 0.6156      & 5.697      \\
SPO          & 21.46     & 0.2671 & 0.2321      & 5.702      \\
\rowcolor{blue!10} SD V1.5+Ours & \underline{\textbf{23.07}}  & \underline{\textbf{0.2755}}   & \underline{\textbf{0.7170}}  & \underline{\textbf{5.831}}  \\ \hline
\end{tabular}}
\vspace{-0.2cm}
\end{table}

\begin{table}[t]
\centering
\small
\renewcommand{\arraystretch}{1.0}
\caption{Comparison of AI feedback on SDXL-based methods.}
\vspace{-0.2cm}
\label{tab:SDXL_Compare_Metric}
\setlength{\tabcolsep}{1.1mm}{
\begin{tabular}{c|cccc}
\hline
Methods             & PickScore & HPSv2  & ImageReward & Aesthetics \\ \hline
SDXL               & 21.91     & 0.2602 & 0.7755      & 5.960      \\ \hline
DNO                & 22.14     & 0.2725 & 0.9053      & 6.042      \\
PromptOpt          & 21.98     & 0.2708 & 0.8671      & 5.881      \\
FreeDom            & 22.13     & 0.2719 & 0.7722      & 5.908      \\ \hline
\rowcolor{blue!10} SDXL+Ours          &  \underline{\textbf{24.90}}       & \underline{\textbf{0.2839}}    &  \underline{\textbf{1.074}}   & \underline{\textbf{6.138}}        \\
Diffusion-DPO      & 22.30     & 0.2741 & 0.9789       & 5.891      \\
\rowcolor{blue!10} Diff-DPO+Ours & \underline{\textbf{24.46}}       & \underline{\textbf{0.2836}}    & \underline{\textbf{1.049}}         & \underline{\textbf{6.116}}        \\
SPO                & 22.81     & 0.2778 & 1.082       & \underline{\textbf{6.319}}      \\
\rowcolor{blue!10} SPO+Ours           & \underline{\textbf{23.85}}       & \underline{\textbf{0.2821}}    & \underline{\textbf{1.166}}         & 6.278        \\ \hline
SD V3.5             & 21.93     & 0.2726 & 0.9697      & 5.775      \\
FLUX               & 22.04     & 0.2760 & 1.011       & 6.077      \\
\hline
\end{tabular}}
\vspace{-0.6cm}
\end{table}

\begin{figure*}[t]
  \centering
  \includegraphics[width=0.80\linewidth]{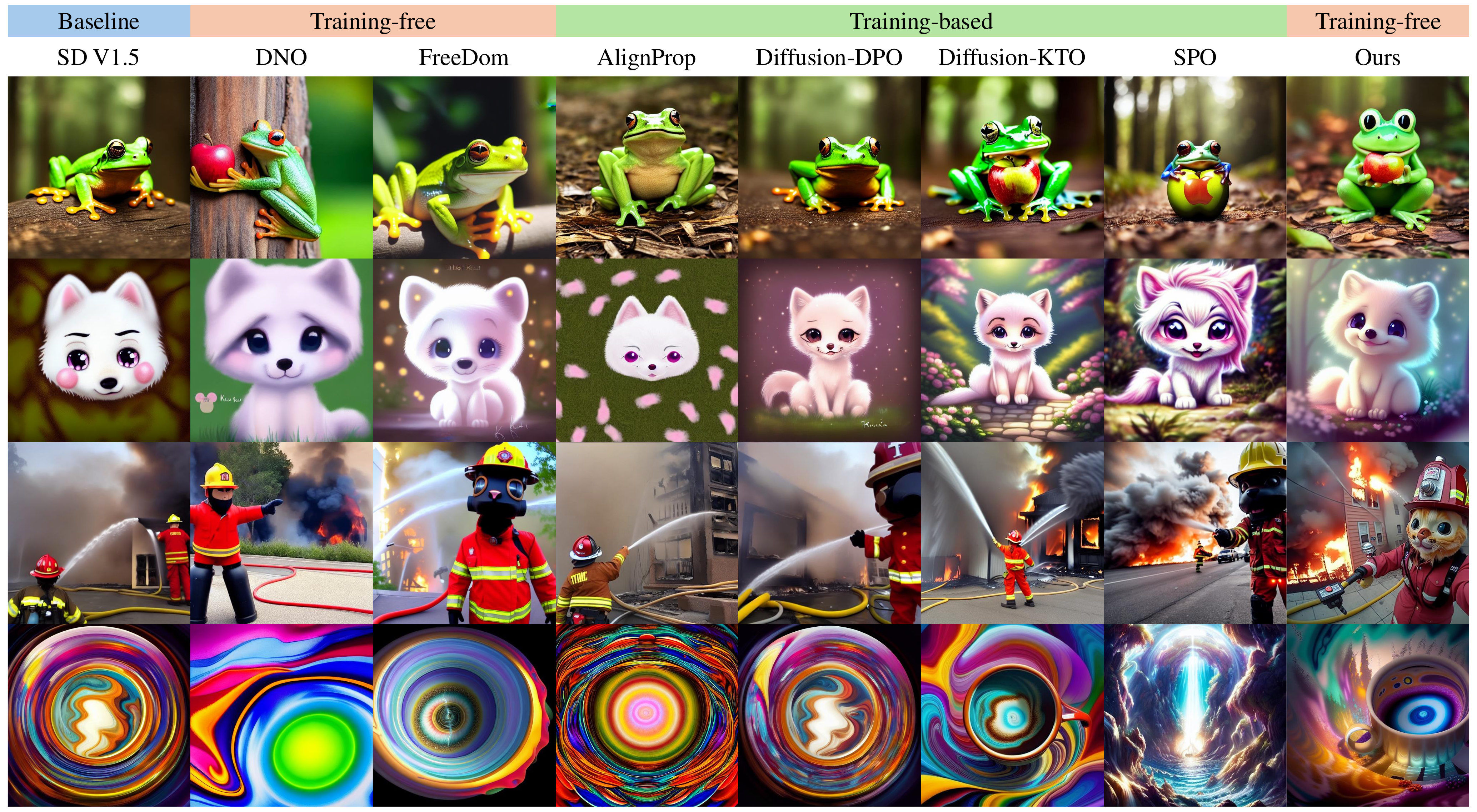}
  \vspace{-0.2cm}
  \caption{Qualitative comparison based on SD V1.5 backbones.}
\label{fig:sd15_compare} 
\vspace{-0.4cm}
\end{figure*}

\section{Experiments}

In this section, different experiments are conducted on a wide range of generative models to verify the effectiveness and flexibility of our method. We validate the performance of our method through comparing with various generative backbones and existing state-of-the-art approaches. We also delve deeper into the role of each component for the guidance during the denoising process, supported by ablation studies for further analysis.

\subsection{Experimental Setting}

\noindent \textbf{Implementation Details.} Our method requires no training and is compatible with various generative models. We employ the GPT-4 \cite{GPT4} as the base LLMs to construct semantic graph. Additionally, we adopt the pre-trained step-aware preference model from SPO \cite{SPO} to facilitate human feedback, developed from PickScore \cite{PickAPic} model and fine-tuned on the human preference pairs dataset, Pick-a-Pic \cite{PickAPic}. All our experiments uses an NVIDIA A100 GPU for SDXL-based methods or V100 GPU for SD V1.5-based methods.

\noindent \textbf{Datasets and Metrics.} We utilized three datasets to validate the effectiveness of our method: 500 unique prompts of the Pick-a-Pic validation set, 500 prompts from HPSv2 benchmark \cite{HPS} and 1000 prompts from Partiprompt \cite{Partiprompt}. We choose four AI feedback models to assess the image quality: PickScore \cite{PickAPic} (general human preference), HPSv2 \cite{HPS} (prompt alignment), ImageReward \cite{ImageReward} (general human preference), Aesthetic Predictor \cite{Aesthetic} (non-text-based visual appeal). \emph{For all metrics, higher values indicate better performance}. All test images are generated using 50 denoising steps during inference.

\subsection{Comparison with Existing Methods}
To verify the effectiveness of our approach, we compare our method with training-based and training-free models. The former comprises human preference learning methods, including AlignProp \cite{AlignProp}, Diffusion-DPO \cite{DiffusionDPO}, Diffusion-KTO \cite{DiffusionKTO}, SPO \cite{AlignProp}. The latter includes DNO \cite{DNO}, PromptOpt \cite{PromptOpt}, FreeDom \cite{FreeDoM}. For SDXL-based frameworks, we make a comparison with the latest models with advanced network architectures (FLUX \cite{FLUX} and SD V3.5 \cite{SD}).

\subsubsection{Quantitative Analysis}
To objectively evaluate the performance of our method, we conduct a quantitative comparison on the Pick-a-Pic dataset, with the results organized in \cref{tab:SD15_Compare_Metric} for SD V1.5-based backbones and \cref{tab:SDXL_Compare_Metric} for SDXL-based backbones, respectively. Under the guidance of our method, models including SD V1.5 \cite{SD}, SDXL \cite{SDXL}, Diffusion-DPO \cite{DiffusionDPO}, and SPO \cite{SPO} achieve notable improvements, surpassing even advanced models like FLUX \cite{FLUX} and SD V3.5 \cite{SD}. Specifically, the training-free methods (DNO \cite{DNO}, PromptOpt \cite{PromptOpt}, FreeDom \cite{FreeDoM}) suffer from underestimation in early steps, resulting in the inferior outcomes. AlignProp \cite{AlignProp} introduces the dependency bias via truncated backpropagation. Diffusion-DPO \cite{DiffusionDPO} and Diffusion-KTO \cite{DiffusionKTO} produce suboptimal results due to trajectory-level preference supervision, causing inaccurate alignment. SPO \cite{SPO} lose the layout and semantic information in early steps. In contrast, our training-free method provides tailored guidance for different denoising steps, generating high-quality images with better layout composition and human preference alignment.

\subsubsection{Qualitative Evaluation}
\begin{figure}[t]
  \centering
  \includegraphics[width=0.94\linewidth]{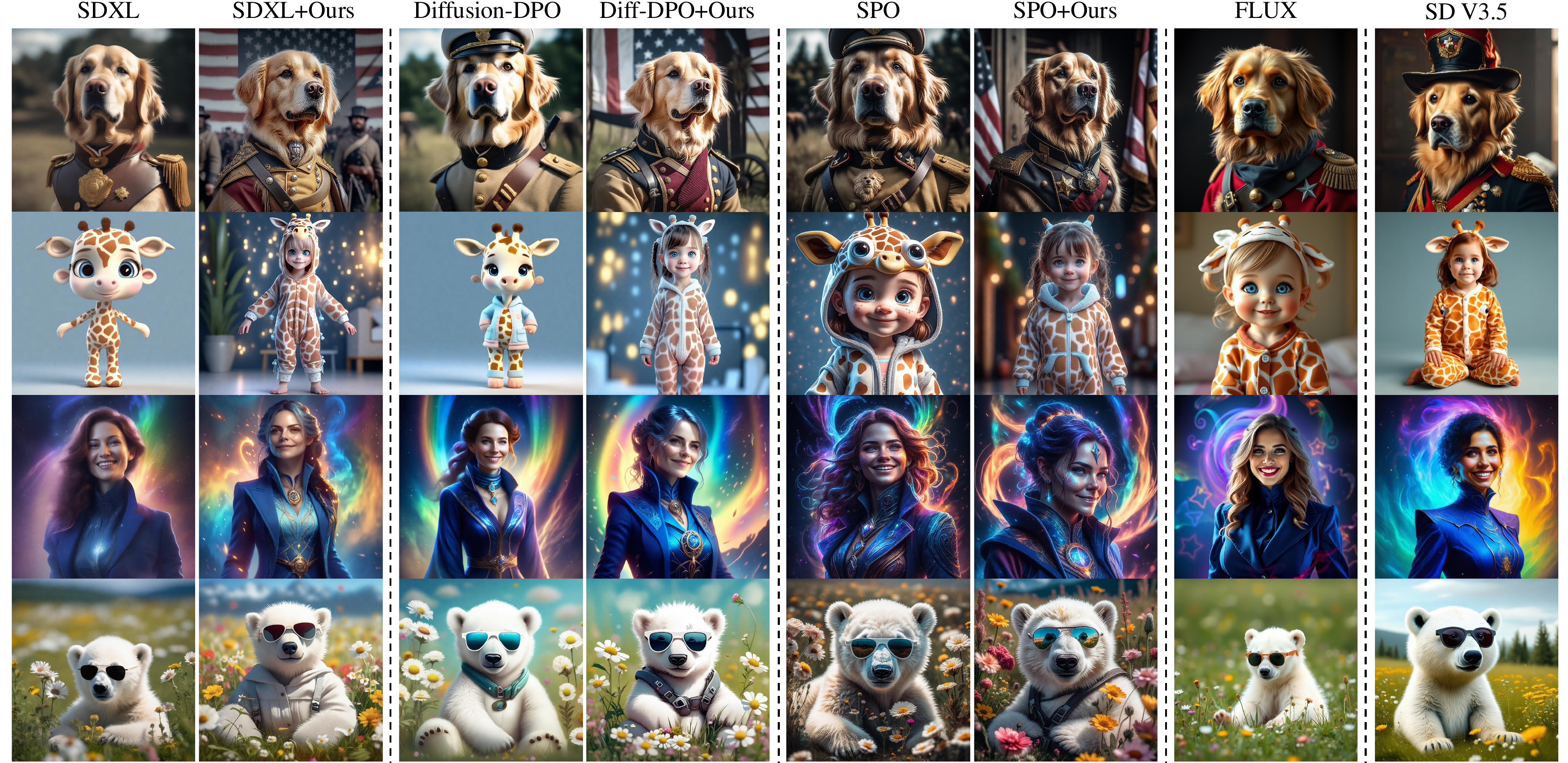}
  \vspace{-0.2cm}
  \caption{Qualitative comparison based on SDXL backbones.}
\label{fig:sdxl_compare} 
\vspace{-0.3cm}
\end{figure}

We compare our method against a baseline model (SD V1.5 \cite{SD}), training-free models (DNO \cite{DNO}, FreeDom \cite{FreeDoM}) and training-based models (AlignProp \cite{AlignProp}, Diffusion-DPO \cite{DiffusionDPO}, Diffusion-KTO \cite{DiffusionKTO}, SPO \cite{SPO}), as illustrated in \cref{fig:sd15_compare}. It is observed that our method contains coherent layout and rich semantic information aligned with the prompt, while achieving overall visual appeal and aesthetic quality. Additionally, we apply our framework to SDXL \cite{SDXL}, Diffusion-DPO \cite{DiffusionDPO}, SPO \cite{SPO} and compare some sample image generation results in \cref{fig:sdxl_compare}. The results show significant improvement in image quality before and after using our method. Specifically, our method enhances the color vibrancy and detail richness, bringing out more defined textures on fur and fabric as well as vivid lighting effects. 

\begin{figure}[t]
  \centering
	\includegraphics[width=0.9\linewidth]{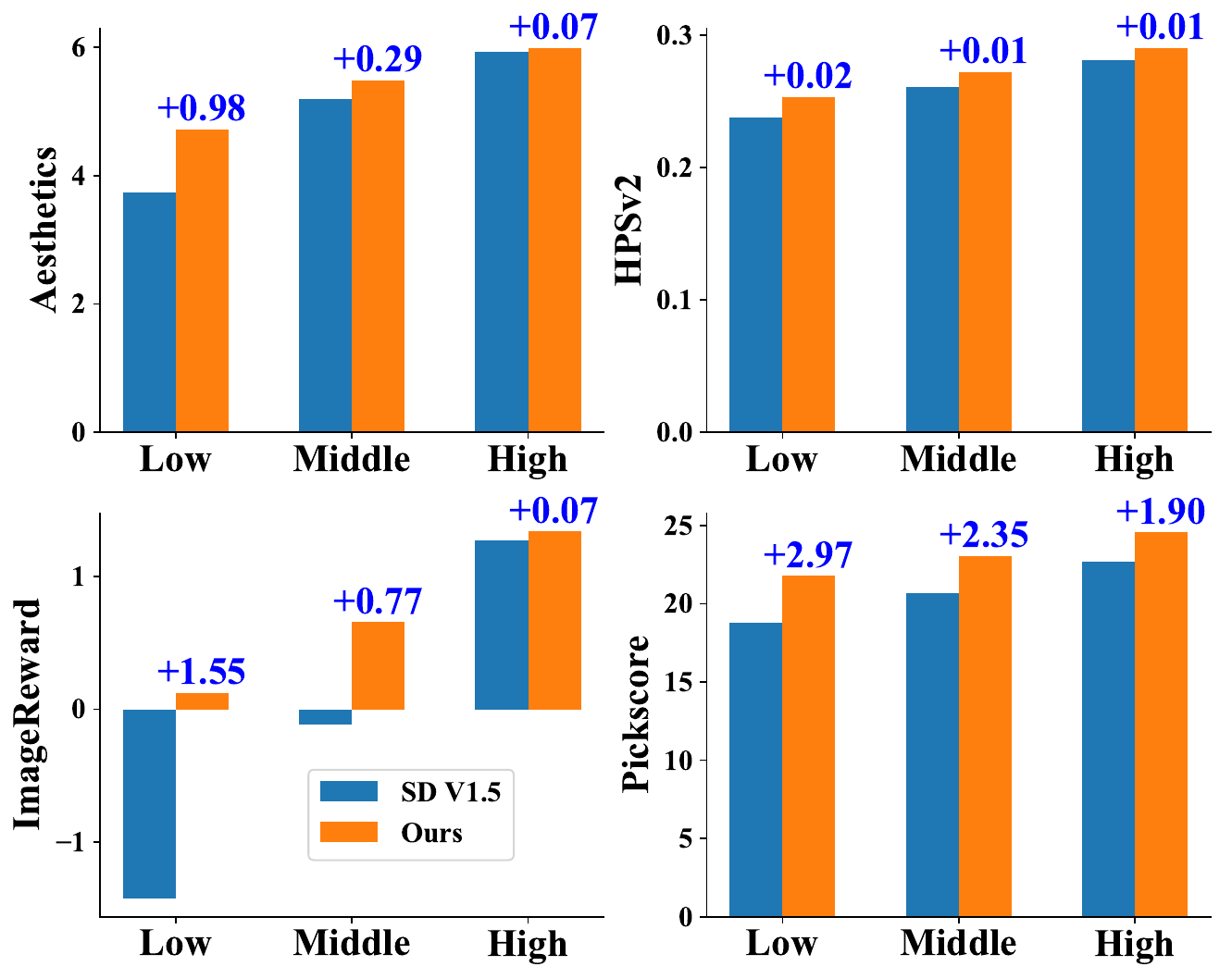}
    \vspace{-0.2cm}
  \caption{The case comparison of improvements between baseline model and  our proposed method.}
  \label{fig:low_middle_high}
  \vspace{-0.5cm}
\end{figure}

\subsubsection{Case Performance}
We compare the performance of our method with baseline model across different cases, as shown in \cref{fig:low_middle_high}. The test dataset is divided into low-, middle-, and high-reward subsets based on the scores achieved in Aesthetics, HPSv2, ImageReward, and PickScore. For aesthetic, our method consistently surpasses the baseline with gains of +0.98, +0.29, and +0.07 across three subsets, respectively. HPSv2 shows slight improvements across all subsets. ImageReward shows a notable +1.55 gain in the low-reward subset, with moderate increases in the other subsets. For PickScore, our method achieves substantial gains of +2.97, +2.35, and +1.90 across the reward levels. These results demonstrate our method boosts the performance on low-reward cases, particularly in layout and visuals, while maintaining competitive results across middle- and high-reward cases.

\begin{figure}[t]
  \centering
    \begin{subfigure}{\linewidth}
    \centering
    \includegraphics[width=0.9\linewidth]{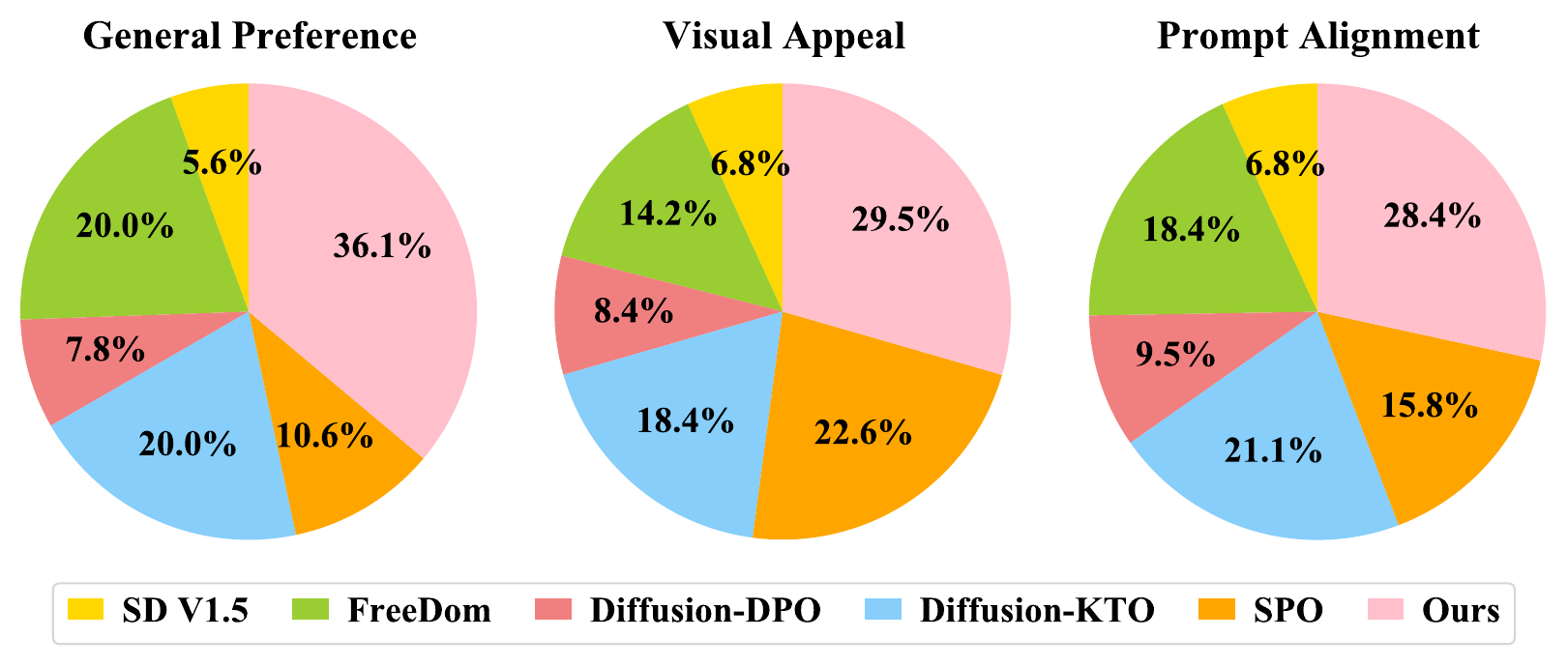}
    \caption{User preference distribution on HPSv2 benchmark.}
    \label{fig:user_study_compare_hps}
  \end{subfigure}
  
  \vspace{0.6em} 

  \begin{subfigure}{\linewidth}
    \centering
    \includegraphics[width=0.86\linewidth]{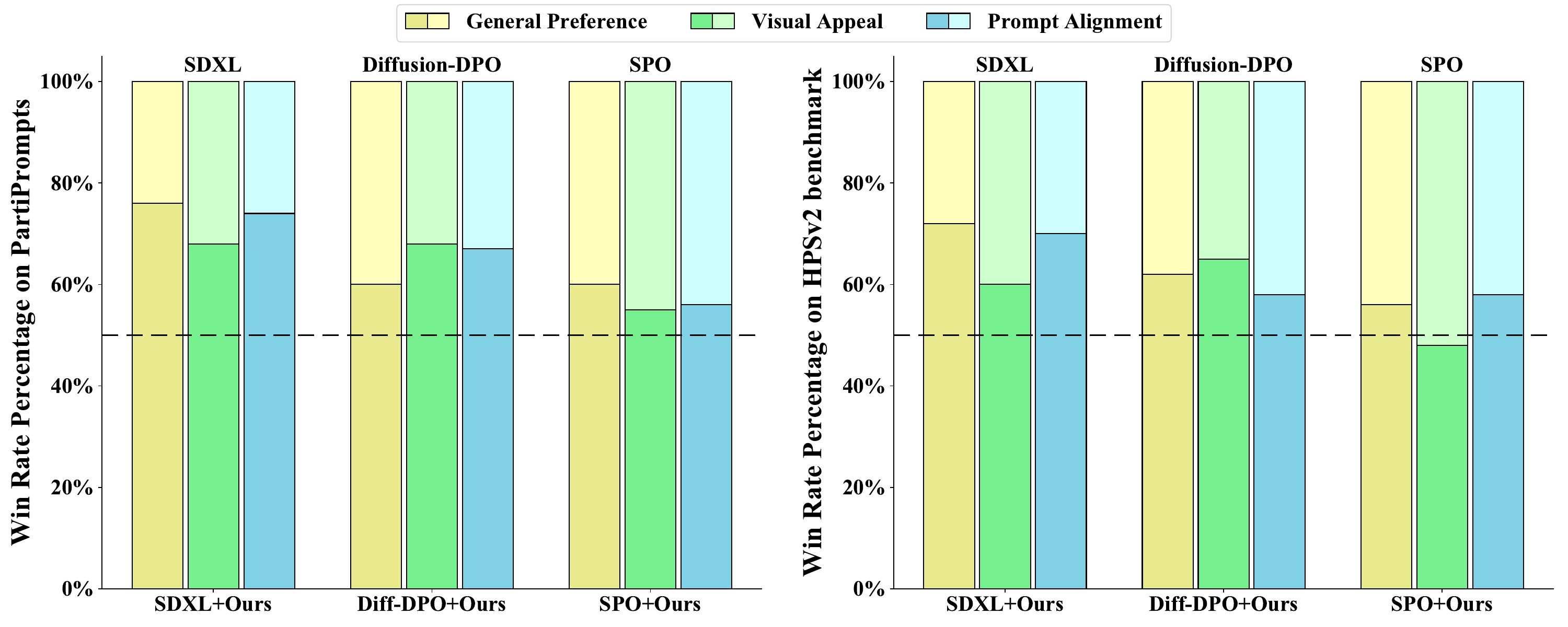}
    \caption{Human evaluation w/ \& w/o our method on PartiPrompts \& HPSv2.}
    \label{fig:user_study_with_and_without}
  \end{subfigure}
  \vspace{-0.7cm} 
  \caption{User study results.}
  \vspace{-0.6cm} 
  \label{fig:user_study}
\end{figure}

\subsubsection{User Study}
We conduct a user study to compare our method (based-on SD V1.5) against SD V1.5 \cite{SD}, FreeDom \cite{FreeDoM}, Diffusion-DPO \cite{DiffusionDPO}, Diffusion-KTO \cite{DiffusionKTO}, SPO \cite{SPO}. We randomly sample 100 unique prompts from HPSv2 \cite{HPS} benchmark and synthesize images with aforementioned methods. We invite 100 participants and request them to compare all generated images from three different aspects: Q1 General Preference (Which image do you prefer given the prompt?), Q2 Visual Appeal (Which image is more visually appealing?), Q3 Prompt Alignment (Which image better fits the text description?). \cref{fig:user_study_compare_hps} shows the approval percentage of each method in three aspects, which demonstrates our method outperforms the previous preference learning models on human feedbacks. Additionally, we conduct a human evaluation on 100 prompts from the PartiPrompts \cite{Partiprompt} dataset and 100 prompts from the HPSv2 \cite{HPS} benchmark, comparing SDXL \cite{SDXL}, Diffusion-DPO \cite{DPO}, and SPO \cite{SPO}, both with and without our method. The win-rate percentage results are reported in the \cref{fig:user_study_with_and_without}, further verifying the effectiveness of our method.

\subsection{Ablation Study}
All ablation experiments are conducted on SD V1.5 \cite{SD} and Pick-a-Pic \cite{PickAPic} dataset if not specified.

\textbf{Effect of semantic alignment loss $\mathcal{L}_{A}$.} The structural formation of image content occurs early in the denoising process, as verified by \cref{tab:SD15_ablation_study}. The inferior performances are due to the lost of semantic components in the prompt.

\textbf{Effect of preference alignment loss $\mathcal{L}_{R}$.} The results in \cref{tab:SD15_ablation_study} highlight the importance of the human feedback. Accurate preference guidance enhances the image-prompt alignment and improves overall image quality. Additionally, we use the PickScore \cite{PickAPic} to replace the step-aware preference model \cite{SPO}, achieving improvements over the baseline model, which further demonstrates the effectiveness and versatility of our method.

\textbf{Effect of adpative weight $w$.} As shown in \cref{tab:SD15_ablation_study}, the adaptive mechanism for balancing these two losses not only preserves essential image structure in the initial steps but also enhances final image fidelity, supporting both content alignment and visual quality.

\begin{table}[t]
\centering
\small
\renewcommand{\arraystretch}{0.9}
\caption{Ablation study results.}
\vspace{-0.25cm}
\label{tab:SD15_ablation_study}
\setlength{\tabcolsep}{1.0mm}{
\begin{tabular}{c|cccc}
\hline
Methods       & PickScore & HPSv2  & ImageReward & Aesthetics \\ \hline
w/o $\mathcal{L}_{A}$      & 22.07     & 0.2546 & 0.6413      & 5.686      \\
w/o $\mathcal{L}_{R}$        & 20.37     & 0.2418 & 0.1230     & 5.426      \\
w/o $w$    & 22.34     & 0.2656 & 0.6748     & 5.708      \\
w/o Polyak step & 20.61     & 0.2640 & 0.2838      & 5.470      \\
w PickSocre & \underline{\textbf{23.38}}     & 0.2746 & 0.5463      & 5.694 \\
Ours (Llama-3.3) & 22.58 & \underline{\textbf{0.2829}} & 0.7048 & 5.802\\
\rowcolor{blue!10} Ours (GPT-4) &  23.07  & 0.2755   & \underline{\textbf{0.7170}}  & \underline{\textbf{5.831}}  \\
\hline
\end{tabular}}
\end{table}

\begin{table}[t]
\centering
\small
\renewcommand{\arraystretch}{0.9}
\caption{Effect of the iteration count in time-travel straregy.}
\vspace{-0.25cm}
\label{tab:SD15_dynamic_recurrent}
\setlength{\tabcolsep}{0.5mm}{
\begin{tabular}{c|ccccc}
\hline
Iteration       & PickScore & HPSv2  & ImageReward & Aesthetics & Avg Time \\ \hline
1      & 20.41     & 0.2423 & 0.6191      & 5.652   & 40s   \\
5        & 21.27     & 0.2610 & 0.6679     & 5.661  & 170s    \\
10    & 22.85     & 0.2677 & 0.7055     & 5.706   & 295s  \\
\rowcolor{blue!10} Dynamic &  \underline{\textbf{23.07}}  & \underline{\textbf{0.2755}}   & \underline{\textbf{0.7170}}  & \underline{\textbf{5.831}}    & 190s \\
\hline
\end{tabular}}
\vspace{-0.5cm}
\end{table}

\textbf{Effect of Polyak step size.} We adopt the Polyak step size \cite{UnderstandImprove} to achieve near-optimal convergence rates for our method. \cref{tab:SD15_ablation_study} demonstrate it is effective to guide the generation process by bridging the substantial gap between unconditional generation and specified condition.

\textbf{Effect of open-source LLMs.} DyMO’s LLM-based text prompt pre-analysis is moderate for LLMs and does not require advanced models (LLM-agnostic). We validate it with open-source Llama3.30-70B (\cref{tab:SD15_ablation_study}), achieving strong performance (22.58)—better than others, slightly below GPT-4-based results (23.07).

\textbf{Effect of dynamic recurrent strategy.} We conduct experiments with fixed iteration counts set to 1, 5, and 10 as presented in \cref{tab:SD15_dynamic_recurrent}. Our approach achieves the better performance with a moderate time cost, demonstrating the effectiveness of dynamic recurrent strategy.

\vspace{-0.2cm}
\section{Conclusion}
\vspace{-0.1cm}
In this work, we propose a novel framework to achieve better diffusion model alignment via dynamic multi-objective scheduling. Apart from text-aware human preference score, we introduce a semantic alignment objective to mitigate the inaccurate estimation for blurred denoised image in early denoising steps. We utilize two objectives in a dynamical coordinated manner to guide model outputs aligned with contextual semantics and human preferences. Extensive experiments show that our algorithm is plug-and-play and enables effective alignment, producing high-quality images with fine layout structure and captivating aesthetics. The performance of our method on different metrics outperforms pre-trained baseline models and surpasses both training-based and training-free approaches.

\clearpage
{
    \small
    \bibliographystyle{ieeenat_fullname}
    \bibliography{MainArxivV2}

\begin{thebibliography}{56}
\providecommand{\natexlab}[1]{#1}
\providecommand{\url}[1]{\texttt{#1}}
\expandafter\ifx\csname urlstyle\endcsname\relax
  \providecommand{\doi}[1]{doi: #1}\else
  \providecommand{\doi}{doi: \begingroup \urlstyle{rm}\Url}\fi

\bibitem[Achiam et~al.(2023)Achiam, Adler, Agarwal, Ahmad, Akkaya, Aleman, Almeida, Altenschmidt, Altman, Anadkat, et~al.]{GPT4}
Josh Achiam, Steven Adler, Sandhini Agarwal, Lama Ahmad, Ilge Akkaya, Florencia~Leoni Aleman, Diogo Almeida, Janko Altenschmidt, Sam Altman, Shyamal Anadkat, et~al.
\newblock Gpt-4 technical report.
\newblock \emph{arXiv preprint arXiv:2303.08774}, 2023.

\bibitem[Bansal et~al.(2023)Bansal, Chu, Schwarzschild, Sengupta, Goldblum, Geiping, and Goldstein]{UGD}
Arpit Bansal, Hong-Min Chu, Avi Schwarzschild, Soumyadip Sengupta, Micah Goldblum, Jonas Geiping, and Tom Goldstein.
\newblock Universal guidance for diffusion models.
\newblock In \emph{Proceedings of the IEEE/CVF Conference on Computer Vision and Pattern Recognition}, pages 843--852, 2023.

\bibitem[Betker et~al.(2023)Betker, Goh, Jing, Brooks, Wang, Li, Ouyang, Zhuang, Lee, Guo, et~al.]{DALLE3}
James Betker, Gabriel Goh, Li Jing, Tim Brooks, Jianfeng Wang, Linjie Li, Long Ouyang, Juntang Zhuang, Joyce Lee, Yufei Guo, et~al.
\newblock Improving image generation with better captions.
\newblock \emph{Computer Science. https://cdn. openai. com/papers/dall-e-3. pdf}, 2\penalty0 (3):\penalty0 8, 2023.

\bibitem[Black et~al.(2023)Black, Janner, Du, Kostrikov, and Levine]{DDPO}
Kevin Black, Michael Janner, Yilun Du, Ilya Kostrikov, and Sergey Levine.
\newblock Training diffusion models with reinforcement learning.
\newblock \emph{arXiv preprint arXiv:2305.13301}, 2023.

\bibitem[Chung et~al.(2023)Chung, Ryu, McCann, Klasky, and Ye]{DPS}
Hyungjin Chung, Dohoon Ryu, Michael~T McCann, Marc~L Klasky, and Jong~Chul Ye.
\newblock Solving 3d inverse problems using pre-trained 2d diffusion models.
\newblock In \emph{Proceedings of the IEEE/CVF Conference on Computer Vision and Pattern Recognition}, pages 22542--22551, 2023.

\bibitem[Clark et~al.(2023)Clark, Vicol, Swersky, and Fleet]{DRaFT}
Kevin Clark, Paul Vicol, Kevin Swersky, and David~J Fleet.
\newblock Directly fine-tuning diffusion models on differentiable rewards.
\newblock \emph{arXiv preprint arXiv:2309.17400}, 2023.

\bibitem[Deckers et~al.(2023)Deckers, Peters, and Potthast]{PromptOpt}
Niklas Deckers, Julia Peters, and Martin Potthast.
\newblock Manipulating embeddings of stable diffusion prompts.
\newblock \emph{arXiv preprint arXiv:2308.12059}, 2023.

\bibitem[Dhariwal and Nichol(2021)]{ClassifierGuidance}
Prafulla Dhariwal and Alexander Nichol.
\newblock Diffusion models beat gans on image synthesis.
\newblock \emph{Advances in neural information processing systems}, 34:\penalty0 8780--8794, 2021.

\bibitem[Dong et~al.(2023)Dong, Xiong, Goyal, Zhang, Chow, Pan, Diao, Zhang, Shum, and Zhang]{RAFT}
Hanze Dong, Wei Xiong, Deepanshu Goyal, Yihan Zhang, Winnie Chow, Rui Pan, Shizhe Diao, Jipeng Zhang, Kashun Shum, and Tong Zhang.
\newblock Raft: Reward ranked finetuning for generative foundation model alignment.
\newblock \emph{arXiv preprint arXiv:2304.06767}, 2023.

\bibitem[Esser et~al.(2024)Esser, Kulal, Blattmann, Entezari, M{\"u}ller, Saini, Levi, Lorenz, Sauer, Boesel, et~al.]{SD3}
Patrick Esser, Sumith Kulal, Andreas Blattmann, Rahim Entezari, Jonas M{\"u}ller, Harry Saini, Yam Levi, Dominik Lorenz, Axel Sauer, Frederic Boesel, et~al.
\newblock Scaling rectified flow transformers for high-resolution image synthesis.
\newblock In \emph{Forty-first International Conference on Machine Learning}, 2024.

\bibitem[Eyring et~al.(2024)Eyring, Karthik, Roth, Dosovitskiy, and Akata]{ReNO}
Luca Eyring, Shyamgopal Karthik, Karsten Roth, Alexey Dosovitskiy, and Zeynep Akata.
\newblock Reno: Enhancing one-step text-to-image models through reward-based noise optimization.
\newblock \emph{arXiv preprint arXiv:2406.04312}, 2024.

\bibitem[Fan et~al.(2024)Fan, Watkins, Du, Liu, Ryu, Boutilier, Abbeel, Ghavamzadeh, Lee, and Lee]{DPOK}
Ying Fan, Olivia Watkins, Yuqing Du, Hao Liu, Moonkyung Ryu, Craig Boutilier, Pieter Abbeel, Mohammad Ghavamzadeh, Kangwook Lee, and Kimin Lee.
\newblock Reinforcement learning for fine-tuning text-to-image diffusion models.
\newblock \emph{Advances in Neural Information Processing Systems}, 36, 2024.

\bibitem[Gambashidze et~al.(2024)Gambashidze, Kulikov, Sosnin, and Makarov]{NCPPO}
Alexander Gambashidze, Anton Kulikov, Yuriy Sosnin, and Ilya Makarov.
\newblock Aligning diffusion models with noise-conditioned perception.
\newblock \emph{arXiv preprint arXiv:2406.17636}, 2024.

\bibitem[Ghosh et~al.(2023)Ghosh, Hajishirzi, and Schmidt]{geneval}
Dhruba Ghosh, Hannaneh Hajishirzi, and Ludwig Schmidt.
\newblock Geneval: An object-focused framework for evaluating text-to-image alignment.
\newblock \emph{Advances in Neural Information Processing Systems}, 36:\penalty0 52132--52152, 2023.

\bibitem[Gu et~al.(2024)Gu, Wang, Yin, Xie, and Zhou]{DiffusionRPO}
Yi Gu, Zhendong Wang, Yueqin Yin, Yujia Xie, and Mingyuan Zhou.
\newblock Diffusion-rpo: Aligning diffusion models through relative preference optimization.
\newblock \emph{arXiv preprint arXiv:2406.06382}, 2024.

\bibitem[Guo et~al.(2024)Guo, Liu, Cui, Li, Yang, and Huang]{InitNO}
Xiefan Guo, Jinlin Liu, Miaomiao Cui, Jiankai Li, Hongyu Yang, and Di Huang.
\newblock Initno: Boosting text-to-image diffusion models via initial noise optimization.
\newblock In \emph{Proceedings of the IEEE/CVF Conference on Computer Vision and Pattern Recognition}, pages 9380--9389, 2024.

\bibitem[Hertz et~al.(2022)Hertz, Mokady, Tenenbaum, Aberman, Pritch, and Cohen-Or]{Prompt2Prompt}
Amir Hertz, Ron Mokady, Jay Tenenbaum, Kfir Aberman, Yael Pritch, and Daniel Cohen-Or.
\newblock Prompt-to-prompt image editing with cross attention control.
\newblock \emph{arXiv preprint arXiv:2208.01626}, 2022.

\bibitem[Ho and Salimans(2022)]{ClassifierFreeGuidance}
Jonathan Ho and Tim Salimans.
\newblock Classifier-free diffusion guidance.
\newblock \emph{arXiv preprint arXiv:2207.12598}, 2022.

\bibitem[Ho et~al.(2020)Ho, Jain, and Abbeel]{DDPM}
Jonathan Ho, Ajay Jain, and Pieter Abbeel.
\newblock Denoising diffusion probabilistic models.
\newblock \emph{Advances in neural information processing systems}, 33:\penalty0 6840--6851, 2020.

\bibitem[Kirstain et~al.(2023)Kirstain, Polyak, Singer, Matiana, Penna, and Levy]{PickAPic}
Yuval Kirstain, Adam Polyak, Uriel Singer, Shahbuland Matiana, Joe Penna, and Omer Levy.
\newblock Pick-a-pic: An open dataset of user preferences for text-to-image generation.
\newblock \emph{Advances in Neural Information Processing Systems}, 36:\penalty0 36652--36663, 2023.

\bibitem[Labs(2024)]{FLUX}
Black~Forest Labs.
\newblock Flux.1-schnell, 2024.
\newblock Accessed: 2024-08-17.

\bibitem[Lee et~al.(2023)Lee, Liu, Ryu, Watkins, Du, Boutilier, Abbeel, Ghavamzadeh, and Gu]{lee2023aligning}
Kimin Lee, Hao Liu, Moonkyung Ryu, Olivia Watkins, Yuqing Du, Craig Boutilier, Pieter Abbeel, Mohammad Ghavamzadeh, and Shixiang~Shane Gu.
\newblock Aligning text-to-image models using human feedback.
\newblock \emph{arXiv preprint arXiv:2302.12192}, 2023.

\bibitem[Li et~al.(2024)Li, Kallidromitis, Gokul, Kato, and Kozuka]{DiffusionKTO}
Shufan Li, Konstantinos Kallidromitis, Akash Gokul, Yusuke Kato, and Kazuki Kozuka.
\newblock Aligning diffusion models by optimizing human utility.
\newblock \emph{arXiv preprint arXiv:2404.04465}, 2024.

\bibitem[Liang et~al.(2024)Liang, Yuan, Gu, Chen, Hang, Li, and Zheng]{SPO}
Zhanhao Liang, Yuhui Yuan, Shuyang Gu, Bohan Chen, Tiankai Hang, Ji Li, and Liang Zheng.
\newblock Step-aware preference optimization: Aligning preference with denoising performance at each step.
\newblock \emph{arXiv preprint arXiv:2406.04314}, 2024.

\bibitem[Liu et~al.(2024)Liu, Shao, Li, Bai, Xiong, Kwok, Helal, and Xie]{Alignment}
Buhua Liu, Shitong Shao, Bao Li, Lichen Bai, Haoyi Xiong, James Kwok, Sumi Helal, and Zeke Xie.
\newblock Alignment of diffusion models: Fundamentals, challenges, and future.
\newblock \emph{arXiv preprint arXiv:2409.07253}, 2024.

\bibitem[Liu et~al.(2023)Liu, Park, Azadi, Zhang, Chopikyan, Hu, Shi, Rohrbach, and Darrell]{MoreControl}
Xihui Liu, Dong~Huk Park, Samaneh Azadi, Gong Zhang, Arman Chopikyan, Yuxiao Hu, Humphrey Shi, Anna Rohrbach, and Trevor Darrell.
\newblock More control for free! image synthesis with semantic diffusion guidance.
\newblock In \emph{Proceedings of the IEEE/CVF Winter Conference on Applications of Computer Vision}, pages 289--299, 2023.

\bibitem[Nair and Patel(2024)]{Dreamguider}
Nithin~Gopalakrishnan Nair and Vishal~M Patel.
\newblock Dreamguider: Improved training free diffusion-based conditional generation.
\newblock \emph{arXiv preprint arXiv:2406.02549}, 2024.

\bibitem[Nichol et~al.(2021)Nichol, Dhariwal, Ramesh, Shyam, Mishkin, McGrew, Sutskever, and Chen]{GLIDE}
Alex Nichol, Prafulla Dhariwal, Aditya Ramesh, Pranav Shyam, Pamela Mishkin, Bob McGrew, Ilya Sutskever, and Mark Chen.
\newblock Glide: Towards photorealistic image generation and editing with text-guided diffusion models.
\newblock \emph{arXiv preprint arXiv:2112.10741}, 2021.

\bibitem[Podell et~al.(2023)Podell, English, Lacey, Blattmann, Dockhorn, M{\"u}ller, Penna, and Rombach]{SDXL}
Dustin Podell, Zion English, Kyle Lacey, Andreas Blattmann, Tim Dockhorn, Jonas M{\"u}ller, Joe Penna, and Robin Rombach.
\newblock Sdxl: Improving latent diffusion models for high-resolution image synthesis.
\newblock \emph{arXiv preprint arXiv:2307.01952}, 2023.

\bibitem[Prabhudesai et~al.(2023)Prabhudesai, Goyal, Pathak, and Fragkiadaki]{AlignProp}
Mihir Prabhudesai, Anirudh Goyal, Deepak Pathak, and Katerina Fragkiadaki.
\newblock Aligning text-to-image diffusion models with reward backpropagation.
\newblock \emph{arXiv preprint arXiv:2310.03739}, 2023.

\bibitem[Rafailov et~al.(2024)Rafailov, Sharma, Mitchell, Manning, Ermon, and Finn]{DPO}
Rafael Rafailov, Archit Sharma, Eric Mitchell, Christopher~D Manning, Stefano Ermon, and Chelsea Finn.
\newblock Direct preference optimization: Your language model is secretly a reward model.
\newblock \emph{Advances in Neural Information Processing Systems}, 36, 2024.

\bibitem[Rombach et~al.(2022)Rombach, Blattmann, Lorenz, Esser, and Ommer]{SD}
Robin Rombach, Andreas Blattmann, Dominik Lorenz, Patrick Esser, and Bj{\"o}rn Ommer.
\newblock High-resolution image synthesis with latent diffusion models.
\newblock In \emph{Proceedings of the IEEE/CVF conference on computer vision and pattern recognition}, pages 10684--10695, 2022.

\bibitem[Saharia et~al.(2022)Saharia, Chan, Saxena, Li, Whang, Denton, Ghasemipour, Gontijo~Lopes, Karagol~Ayan, Salimans, et~al.]{imagen}
Chitwan Saharia, William Chan, Saurabh Saxena, Lala Li, Jay Whang, Emily~L Denton, Kamyar Ghasemipour, Raphael Gontijo~Lopes, Burcu Karagol~Ayan, Tim Salimans, et~al.
\newblock Photorealistic text-to-image diffusion models with deep language understanding.
\newblock \emph{Advances in neural information processing systems}, 35:\penalty0 36479--36494, 2022.

\bibitem[Sauer et~al.(2024)Sauer, Boesel, Dockhorn, Blattmann, Esser, and Rombach]{SD3Turbo}
Axel Sauer, Frederic Boesel, Tim Dockhorn, Andreas Blattmann, Patrick Esser, and Robin Rombach.
\newblock Fast high-resolution image synthesis with latent adversarial diffusion distillation.
\newblock \emph{arXiv preprint arXiv:2403.12015}, 2024.

\bibitem[Schuhmann(2022)]{Aesthetic}
Christoph Schuhmann.
\newblock Laion-aesthetics.
\newblock \url{https://laion.ai/blog/laion-aesthetics/}, 2022.
\newblock Accessed: 2023 - 11- 10.

\bibitem[Shen et~al.(2024)Shen, Jiang, Wang, Yang, Han, and Li]{UnderstandImprove}
Yifei Shen, Xinyang Jiang, Yezhen Wang, Yifan Yang, Dongqi Han, and Dongsheng Li.
\newblock Understanding and improving training-free loss-based diffusion guidance, 2024.

\bibitem[Song et~al.(2023)Song, Vahdat, Mardani, and Kautz]{IIGDM}
Jiaming Song, Arash Vahdat, Morteza Mardani, and Jan Kautz.
\newblock Pseudoinverse-guided diffusion models for inverse problems.
\newblock In \emph{International Conference on Learning Representations}, 2023.

\bibitem[Song and Ermon(2019)]{song2019generative}
Yang Song and Stefano Ermon.
\newblock Generative modeling by estimating gradients of the data distribution.
\newblock \emph{Advances in neural information processing systems}, 32, 2019.

\bibitem[Song et~al.(2020)Song, Sohl-Dickstein, Kingma, Kumar, Ermon, and Poole]{SDE}
Yang Song, Jascha Sohl-Dickstein, Diederik~P Kingma, Abhishek Kumar, Stefano Ermon, and Ben Poole.
\newblock Score-based generative modeling through stochastic differential equations.
\newblock \emph{arXiv preprint arXiv:2011.13456}, 2020.

\bibitem[Sun et~al.(2024)Sun, Wang, Tan, Dong, Ma, Li, and Gong]{EGGen}
Zhenhong Sun, Junyan Wang, Zhiyu Tan, Daoyi Dong, Hailan Ma, Hao Li, and Dong Gong.
\newblock Eggen: Image generation with multi-entity prior learning through entity guidance.
\newblock In \emph{Proceedings of the 32nd ACM International Conference on Multimedia}, pages 6637--6645, 2024.

\bibitem[Tang et~al.(2024)Tang, Peng, Tang, Hong, Wang, and Chang]{DNO}
Zhiwei Tang, Jiangweizhi Peng, Jiasheng Tang, Mingyi Hong, Fan Wang, and Tsung-Hui Chang.
\newblock Tuning-free alignment of diffusion models with direct noise optimization.
\newblock \emph{arXiv preprint arXiv:2405.18881}, 2024.

\bibitem[Touvron et~al.(2023)Touvron, Lavril, Izacard, Martinet, Lachaux, Lacroix, Rozi{\`e}re, Goyal, Hambro, Azhar, et~al.]{Llama}
Hugo Touvron, Thibaut Lavril, Gautier Izacard, Xavier Martinet, Marie-Anne Lachaux, Timoth{\'e}e Lacroix, Baptiste Rozi{\`e}re, Naman Goyal, Eric Hambro, Faisal Azhar, et~al.
\newblock Llama: Open and efficient foundation language models.
\newblock \emph{arXiv preprint arXiv:2302.13971}, 2023.

\bibitem[Tumanyan et~al.(2023)Tumanyan, Geyer, Bagon, and Dekel]{P2P}
Narek Tumanyan, Michal Geyer, Shai Bagon, and Tali Dekel.
\newblock Plug-and-play diffusion features for text-driven image-to-image translation.
\newblock In \emph{Proceedings of the IEEE/CVF Conference on Computer Vision and Pattern Recognition}, pages 1921--1930, 2023.

\bibitem[Wallace et~al.(2023)Wallace, Gokul, Ermon, and Naik]{DOODL}
Bram Wallace, Akash Gokul, Stefano Ermon, and Nikhil Naik.
\newblock End-to-end diffusion latent optimization improves classifier guidance.
\newblock In \emph{Proceedings of the IEEE/CVF International Conference on Computer Vision}, pages 7280--7290, 2023.

\bibitem[Wallace et~al.(2024)Wallace, Dang, Rafailov, Zhou, Lou, Purushwalkam, Ermon, Xiong, Joty, and Naik]{DiffusionDPO}
Bram Wallace, Meihua Dang, Rafael Rafailov, Linqi Zhou, Aaron Lou, Senthil Purushwalkam, Stefano Ermon, Caiming Xiong, Shafiq Joty, and Nikhil Naik.
\newblock Diffusion model alignment using direct preference optimization.
\newblock In \emph{Proceedings of the IEEE/CVF Conference on Computer Vision and Pattern Recognition}, pages 8228--8238, 2024.

\bibitem[Wang et~al.(2024)Wang, Yu, Luo, Fan, and Zhang]{wang2024magic}
Hao Wang, Yongsheng Yu, Tiejian Luo, Heng Fan, and Libo Zhang.
\newblock Magic: Multi-modality guided image completion.
\newblock In \emph{The Twelfth International Conference on Learning Representations}, 2024.

\bibitem[Wu et~al.(2023)Wu, Hao, Sun, Chen, Zhu, Zhao, and Li]{HPS}
Xiaoshi Wu, Yiming Hao, Keqiang Sun, Yixiong Chen, Feng Zhu, Rui Zhao, and Hongsheng Li.
\newblock Human preference score v2: A solid benchmark for evaluating human preferences of text-to-image synthesis.
\newblock \emph{arXiv preprint arXiv:2306.09341}, 2023.

\bibitem[Wu et~al.(2024)Wu, Hao, Zhang, Sun, Huang, Song, Liu, and Li]{DRTune}
Xiaoshi Wu, Yiming Hao, Manyuan Zhang, Keqiang Sun, Zhaoyang Huang, Guanglu Song, Yu Liu, and Hongsheng Li.
\newblock Deep reward supervisions for tuning text-to-image diffusion models.
\newblock \emph{arXiv preprint arXiv:2405.00760}, 2024.

\bibitem[Xu et~al.(2024)Xu, Liu, Wu, Tong, Li, Ding, Tang, and Dong]{ImageReward}
Jiazheng Xu, Xiao Liu, Yuchen Wu, Yuxuan Tong, Qinkai Li, Ming Ding, Jie Tang, and Yuxiao Dong.
\newblock Imagereward: Learning and evaluating human preferences for text-to-image generation.
\newblock \emph{Advances in Neural Information Processing Systems}, 36, 2024.

\bibitem[Yang et~al.(2024{\natexlab{a}})Yang, Tao, Lyu, Ge, Chen, Shen, Zhu, and Li]{D3PO}
Kai Yang, Jian Tao, Jiafei Lyu, Chunjiang Ge, Jiaxin Chen, Weihan Shen, Xiaolong Zhu, and Xiu Li.
\newblock Using human feedback to fine-tune diffusion models without any reward model.
\newblock In \emph{Proceedings of the IEEE/CVF Conference on Computer Vision and Pattern Recognition}, pages 8941--8951, 2024{\natexlab{a}}.

\bibitem[Yang et~al.(2024{\natexlab{b}})Yang, Chen, and Zhou]{DenseReward}
Shentao Yang, Tianqi Chen, and Mingyuan Zhou.
\newblock A dense reward view on aligning text-to-image diffusion with preference.
\newblock \emph{arXiv preprint arXiv:2402.08265}, 2024{\natexlab{b}}.

\bibitem[Ye et~al.(2024)Ye, Lin, Han, Xu, Liu, Liang, Ma, Zou, and Ermon]{TFG}
Haotian Ye, Haowei Lin, Jiaqi Han, Minkai Xu, Sheng Liu, Yitao Liang, Jianzhu Ma, James Zou, and Stefano Ermon.
\newblock Tfg: Unified training-free guidance for diffusion models.
\newblock \emph{arXiv preprint arXiv:2409.15761}, 2024.

\bibitem[Yu et~al.(2022)Yu, Xu, Koh, Luong, Baid, Wang, Vasudevan, Ku, Yang, Ayan, et~al.]{Partiprompt}
Jiahui Yu, Yuanzhong Xu, Jing~Yu Koh, Thang Luong, Gunjan Baid, Zirui Wang, Vijay Vasudevan, Alexander Ku, Yinfei Yang, Burcu~Karagol Ayan, et~al.
\newblock Scaling autoregressive models for content-rich text-to-image generation.
\newblock \emph{arXiv preprint arXiv:2206.10789}, 2\penalty0 (3):\penalty0 5, 2022.

\bibitem[Yu et~al.(2023)Yu, Wang, Zhao, Ghanem, and Zhang]{FreeDoM}
Jiwen Yu, Yinhuai Wang, Chen Zhao, Bernard Ghanem, and Jian Zhang.
\newblock Freedom: Training-free energy-guided conditional diffusion model.
\newblock In \emph{Proceedings of the IEEE/CVF International Conference on Computer Vision}, pages 23174--23184, 2023.

\bibitem[Zhang et~al.(2025)Zhang, Yu, and Wu]{EBAMA}
Yasi Zhang, Peiyu Yu, and Ying~Nian Wu.
\newblock Object-conditioned energy-based attention map alignment in text-to-image diffusion models.
\newblock In \emph{European Conference on Computer Vision}, pages 55--71. Springer, 2025.

\bibitem[Zhao et~al.(2020)Zhao, Liu, Lin, Zhu, and Han]{RandomAugmentation}
Shengyu Zhao, Zhijian Liu, Ji Lin, Jun-Yan Zhu, and Song Han.
\newblock Differentiable augmentation for data-efficient gan training.
\newblock \emph{Advances in neural information processing systems}, 33:\penalty0 7559--7570, 2020.

\end{thebibliography}
}

\clearpage
\setcounter{page}{1}
\maketitlesupplementary

\section{More Details of the Method}

\subsection{More Details of Dynamic Scheduling in DyMO}
\begin{figure*}[t]
\centering
 \includegraphics[width=1\linewidth]{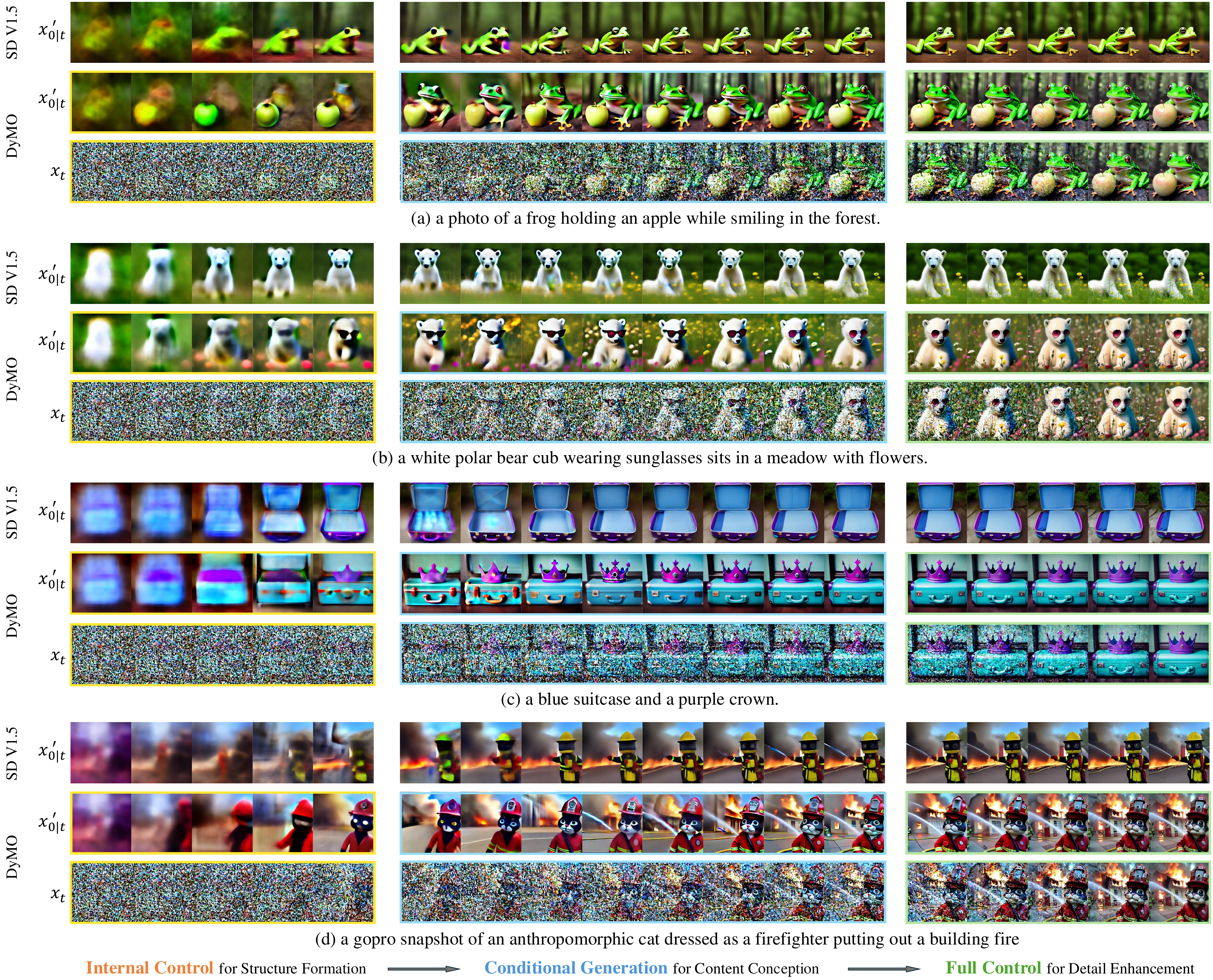}
\vspace{-1em}
\caption{The entire denoising process of SD V1.5 and DyMO, where $\mathbf{x}_{t}$ and $\mathbf{x}_{0|t}^{\prime}$ denote the noisy images and one-step predicted clean images at step $t$, respectively.}
\label{fig:demonstration_case}
\end{figure*}

The proposed method includes two alignment objectives (as guidance in inference) and dynamically schedules the usage of the objective (\cref{subsec:multi_guidance}), the step size, and the time-travel recurrent steps (\cref{subsec:dynamic_recurrent}). We demonstrate the generative denoising process in \cref{fig:demonstration_case} with examples. It shows the intermediate images of the baseline model and the proposed method (including the noisy images and the predicted clean images). 

\par
To achieve alignment in the generative denoising process, we used the text-aware human preference score $\mathcal{L}_{R}(\mathbf{x}_{0|t}^{\prime}, \mathbf{c}, t)$ to guide the denoising process with the gradient computed for the intermediate noisy images.
As discussed in the main paper, the samples in the early stage are highly noisy and the predicted clean images obtained through one-step approximation are also blurred, as shown in \cref{fig:demonstration_case}, demonstrating that the preference model cannot provide effective and accurate guidance. Especially, semantic context is often established during the initial denoising steps, yet it lacks effective supervision at this critical stage. Relying on the proposed semantic alignment objective $\mathcal{L}_{A}$ depending on the semantic contents reflected in the text-vision attention maps, the proposed method can effectively guide the alignment in the early noisy steps. Some previous works also control the contents by relying on the manipulation on the noise map or attention maps, which are restricted to pre-defined and additionally given layout \cite{EGGen, FreeDoM} or oversimplified the semantics \cite{EBAMA}, limiting them to simple text prompts and restricted usage cases. 

\par
We integrate the two objectives $\mathcal{L}_{R}$ and $\mathcal{L}_{A}$ into a dynamic scheduling process through the weight $w$ (and the corresponding $1-w$) in \cref{eq:dy_w}, which are also represented as adaptive weights $w_A$ and $w_R$ for convenience in \cref{algo:code}. 
The weights are step-$t$-dependent and are adjusted according to the relative changes of $\mathbf{z}$. Considering the strength and the different goals of the two objectives, we let $\mathcal{L}_{R}$ and $\mathcal{L}_{A}$ be functional more at the early and later stages, respectively, as shown in \cref{algo:code}. 
Additionally, we also dynamically schedule the step size of the update of $\mathbf{z}$ and the number of recurrent steps in the dynamic time-travel strategy, as introduced in \cref{subsec:dynamic_recurrent}. The dynamic adjustment operations enable the model to achieve an adaptive alignment process that can be aware of the status of specific steps, leading to both better effectiveness and efficiency.

\section{Additional Experimental Details and Results}

\subsection{Details of Prompts used in Experiments}

The text prompts used to generate the images in \cref{fig:first_image}, \cref{fig:sd15_compare}, and \cref{fig:sdxl_compare} are summarized in \cref{tab:first_image_prompt_list}, \cref{tab:sd15_compare_prompt_list}, and \cref{tab:sdxl_compare_prompt_list}, respectively, providing a clear reference for the input descriptions corresponding to each figure.

\subsection{Constructing Semantic Graph from Input Text Prompts}
In \cref{subsec:semantic_alignment}, we introduce semantic alignment guidance for more effective alignment of the image contents with the semantic contents and intention in the user's input text prompts, which requires extracting the semantic information (and semantic graph) from the text. It is achieved through a pre-trained large language model (LLM) with some designed instruction prompts, as mentioned in the paper. We provide some exemplar cases of text semantic graph in \cref{fig:graph_example}.

\begin{figure}[t]
\centering
 \includegraphics[width=1\linewidth]{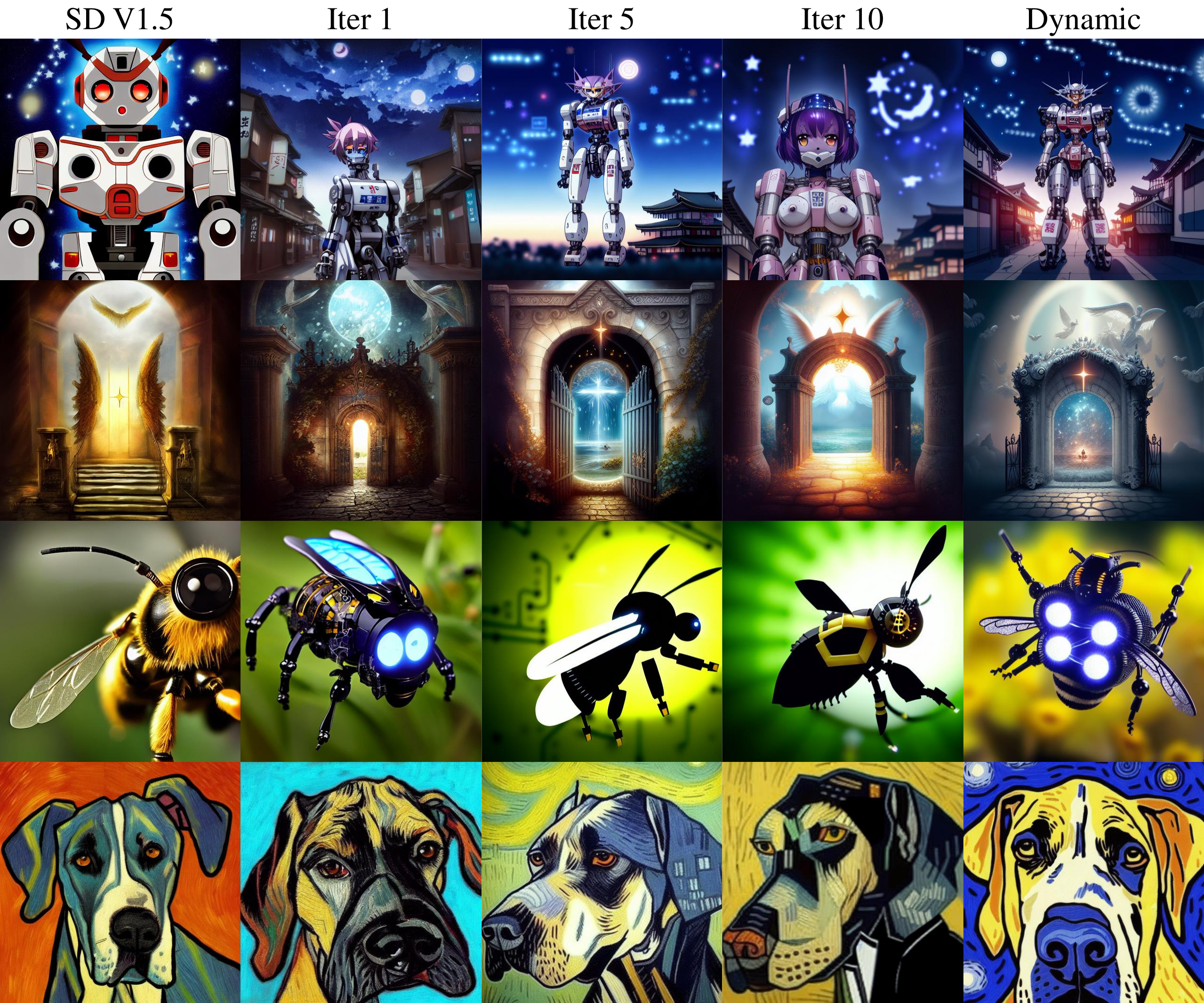}
\caption{Qualitative comparison of iteration count in time-travel straregy. The prompts from top to bottom are: (1) A portrait of an anime mecha robot with a Japanese town background and a starred night sky. (2) The gate to the eternal kingdom of angels, fantasy, digital painting, HD, detailed. (3) mechanical bee flying in nature, electronics, motors, wires, buttons, lcd, led instead of eyes, antennas instead of feet. (4) A Great dane dog in the style of Vincent Van Gogh.}
\label{fig:iter_ablation}
\end{figure}

\subsection{Additional Qualitative Results}
We provide more visual results for qualitative evaluation. 

\par
\begin{figure*}[t]
\centering
 \includegraphics[width=1\linewidth]{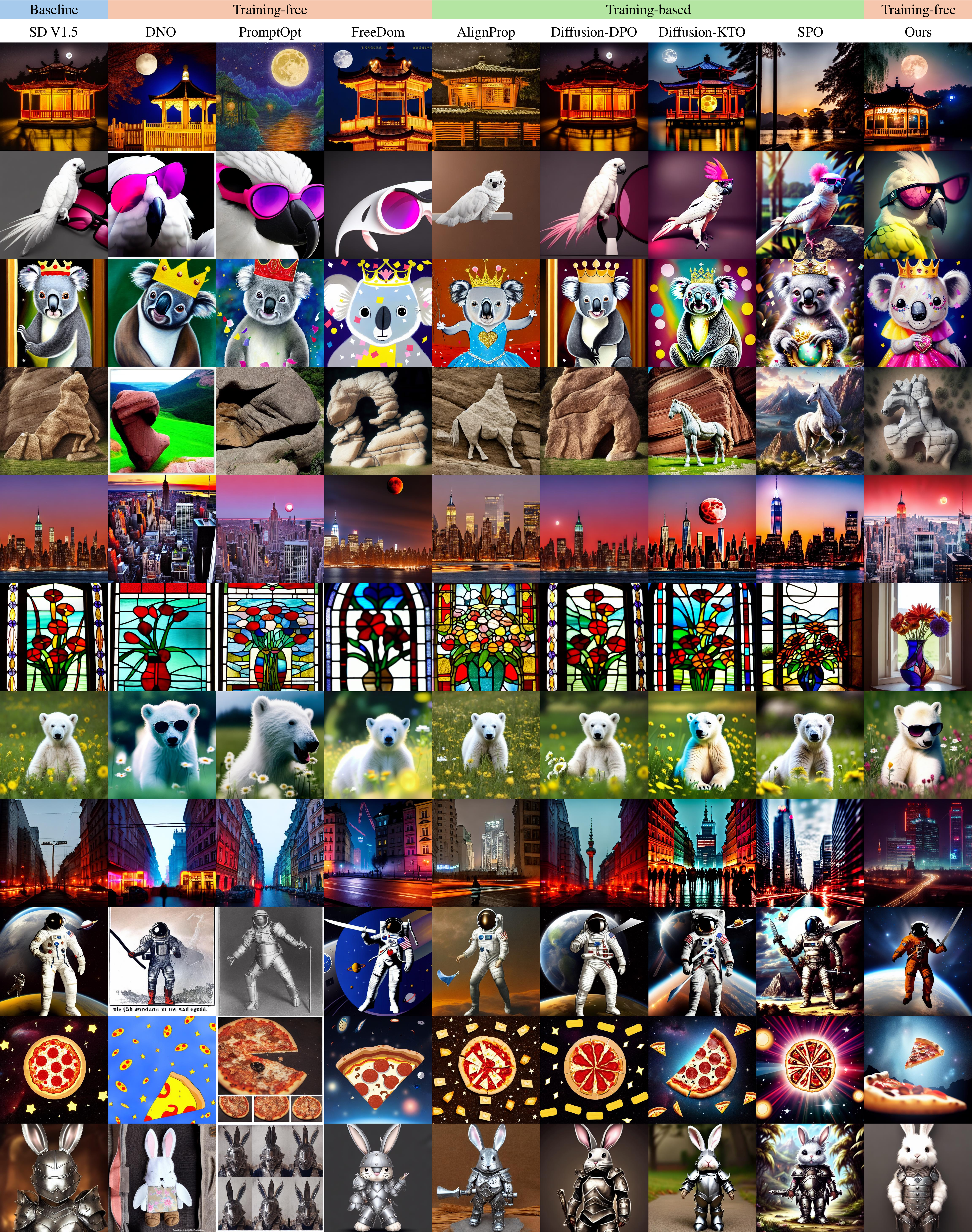}
\vspace{-1em}
\caption{Qualitative comparison based on SD V1.5 backbones. The prompts are provided in the \cref{tab:sd15_compare_more_prompt_list}.}
\label{fig:compare_sd15_more_our}
\end{figure*}

In \cref{fig:sd15_compare} in the main paper, we provide some exemplar cases of comparing our methods with the baseline model (SD V1.5 \cite{SD}), training-free models (\eg, DNO \cite{DNO}, PromptOpt \cite{PromptOpt}, FreeDom \cite{FreeDoM}) and training-based models (\eg, AlignProp \cite{AlignProp}, Diffusion-DPO \cite{DiffusionDPO}, Diffusion-KTO \cite{DiffusionKTO}, SPO \cite{SPO}). We provide a more comprehensive comparison with more examples of generated images in \cref{fig:compare_sd15_more_our} with the corresponding text prompts in \cref{tab:sd15_compare_more_prompt_list}. In \cref{fig:compare_sd15_more_our} and \cref{fig:sd15_compare}, all the compared methods are based on the same baseline model SD V1.5. It is observed that our method contains semantic information aligning better with the input prompts, 
such as the frog holding an apple in \cref{fig:sd15_compare} as well as the horse shape in the 4th row and the sunglass in the 7th row in \cref{fig:compare_sd15_more_our}. The generated images by our methods contain more visually appealing appearances (\eg, rich details and visual characteristics) that are highly aligned with human preferences, such as color vibrancy (2nd and 3rd rows in \cref{fig:compare_sd15_more_our}), vivid lighting effects (5th and 8th rows in \cref{fig:compare_sd15_more_our}) and detailed textures (1st, 6th and last rows in \cref{fig:compare_sd15_more_our}), etc. 

\begin{table*}[t]
\centering
\renewcommand{\arraystretch}{1.2}
\caption{Geneval Benchmark evaluation based on SD V1.5.}
\label{tab:SD15_Compare_Geneval}
\setlength{\tabcolsep}{3.0mm}{
\begin{tabular}{c|ccccccc}
\hline
Methods       & Overall & Single object & Two object & Counting & Colors & Position &  Color attribution
 \\ \hline
SD V1.5      & 0.42     & 1.00 & 0.38      & 0.35  & 0.77 & 0.04 & 0.00    \\ \hline
DNO          & 0.43     & 0.96 & 0.35     & 0.35  & 0.82 & 0.04 & 0.05    \\
PromptOpt    & 0.39     & 0.96 & 0.25     & 0.23  & 0.80 & 0.04 & 0.05    \\
FreeDom      & 0.52     & 1.00 & 0.56      & \underline{\textbf{0.64}}  & 0.80 & 0.02 & 0.13    \\ \hline
AlignProp    & 0.42     & 1.00 & 0.23      & 0.41  & 0.74 & 0.04 & 0.06    \\
Diffusion-DPO & 0.47     & 1.00 & 0.48      & 0.47  & 0.80 & 0.02 & 0.05    \\
Diffusion-KTO & 0.49     & 1.00 & 0.53      & 0.41  & 0.82 & 0.06 & 0.11    \\
SPO          & 0.47     & 0.96 & 0.48      & 0.47   & 0.74 & \underline{\textbf{0.11}} & 0.05   \\
\rowcolor{blue!10} SD V1.5+Ours & \underline{\textbf{0.57}}  & \textbf{1.00}   & \underline{\textbf{0.72}}  & 0.47 & \underline{\textbf{0.83}} & 0.07 & \underline{\textbf{0.34}}  \\ \hline
\end{tabular}}
\end{table*}

\par
\begin{figure*}[t]
\centering
 \includegraphics[width=1\linewidth]{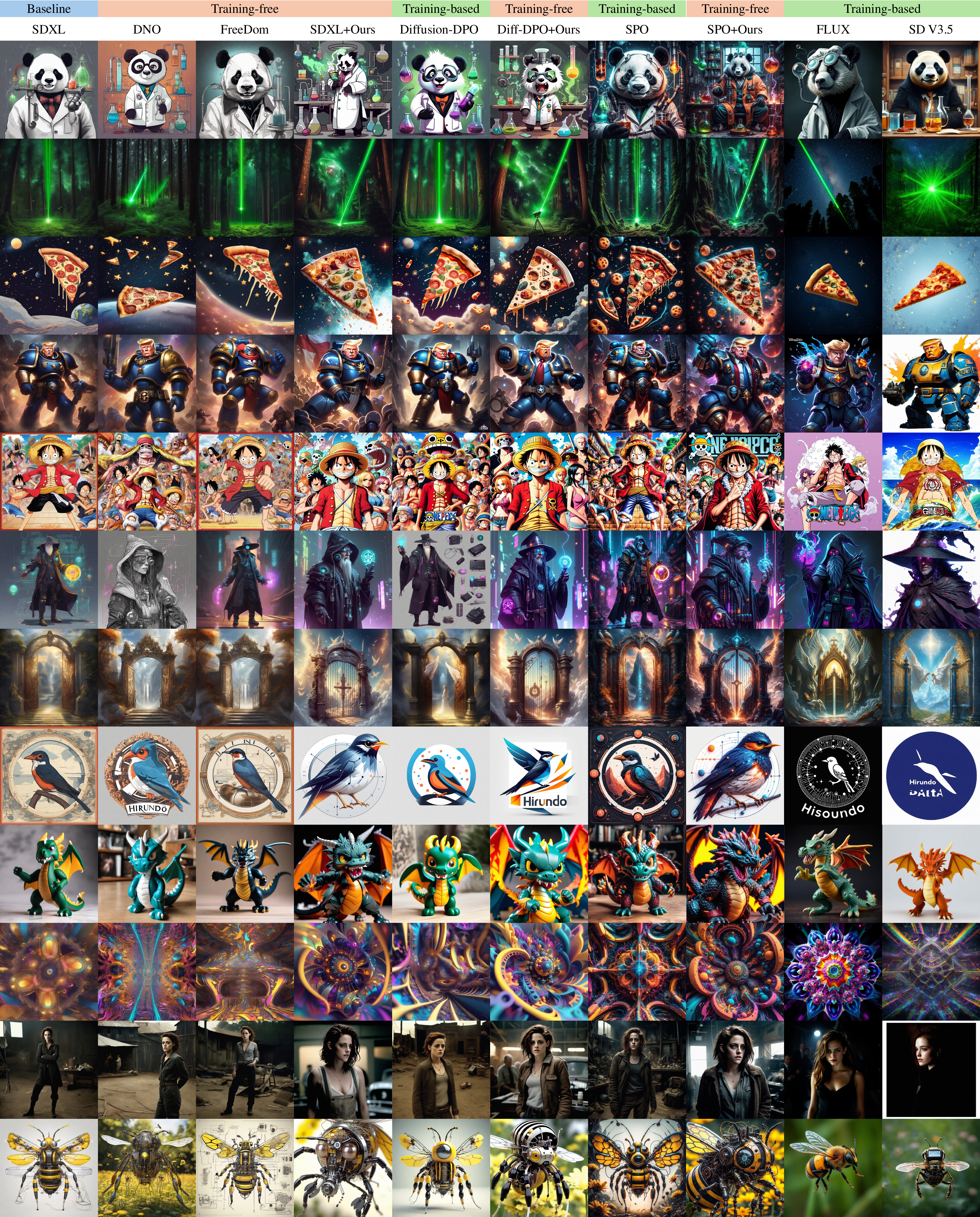}
\vspace{-1em}
\caption{Qualitative comparison based on SDXL backbones. The prompts are provided in the \cref{tab:sdxl_compare_more_prompt_list}.}
\label{fig:compare_sdxl_more1}
\end{figure*}
\begin{figure*}[t]
\centering
 \includegraphics[width=1\linewidth]{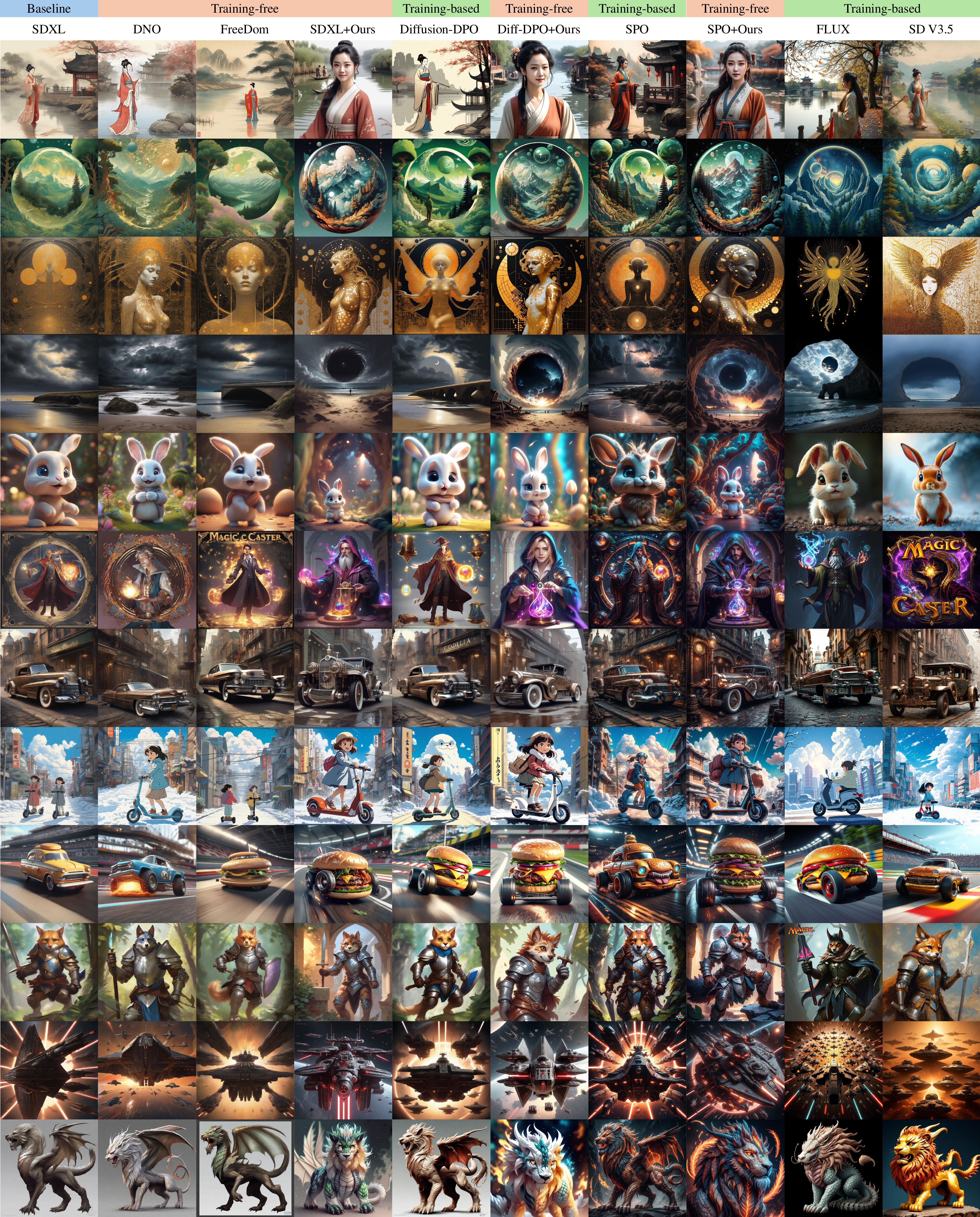}
\vspace{-1em}
\caption{Qualitative comparison based on SDXL backbones. The prompts are provided in the \cref{tab:sdxl_compare_more_prompt_list}.}
\label{fig:compare_sdxl_more2}
\end{figure*}

Beyond the baseline SD V1.5, we also validate our method by applying it to other models such as SDXL \cite{SDXL}, Diffusion-DPO \cite{DiffusionDPO}, and SPO \cite{SPO}, and demonstrate the results in Fig. \cref{fig:compare_sdxl_more1} and \cref{fig:compare_sdxl_more2} (with the corresponding text prompts in \cref{tab:sdxl_compare_more_prompt_list}). \cref{fig:sdxl_compare} (with text in \cref{tab:sdxl_compare_prompt_list}) in the main paper also includes a small set of results based on SDXL. And \cref{tab:SD15_Compare_Metric} demonstrates the numerical results of the comparison. The results show that the proposed training-free alignment method can generally improve the performance after adding it to different pre-trained models. Compared to the baseline models, our approach generates high-quality images more closely aligned with contextual semantics and better cater to human preferences. We also show the latest 
state-of-the-art generative models (FLUX \cite{FLUX}, SD V3.5 \cite{SD3}) as a reference. The effectiveness of the proposed method DyMO is still obvious in the comparison with them, in terms of visual coherence and detail fidelity.

\subsection{Additional Quantitative Results}
We conduct quantitative evaluation on Geneval benchmark \cite{geneval} and show comparisons in \cref{tab:SD15_Compare_Geneval}. Our method performs very well and shows superiority in many aspects, \eg, overall, attribute binding and object synthesis. The proposed semantic alignment can help DyMO on multi-object synthesis. \cref{tab:SD15_Compare_Geneval} also shows that DyMO performs well in multiple objects and counting.

\subsection{Additional Details on Human Evaluation}
\begin{figure*}[t]
\centering
 \includegraphics[width=0.9\linewidth]{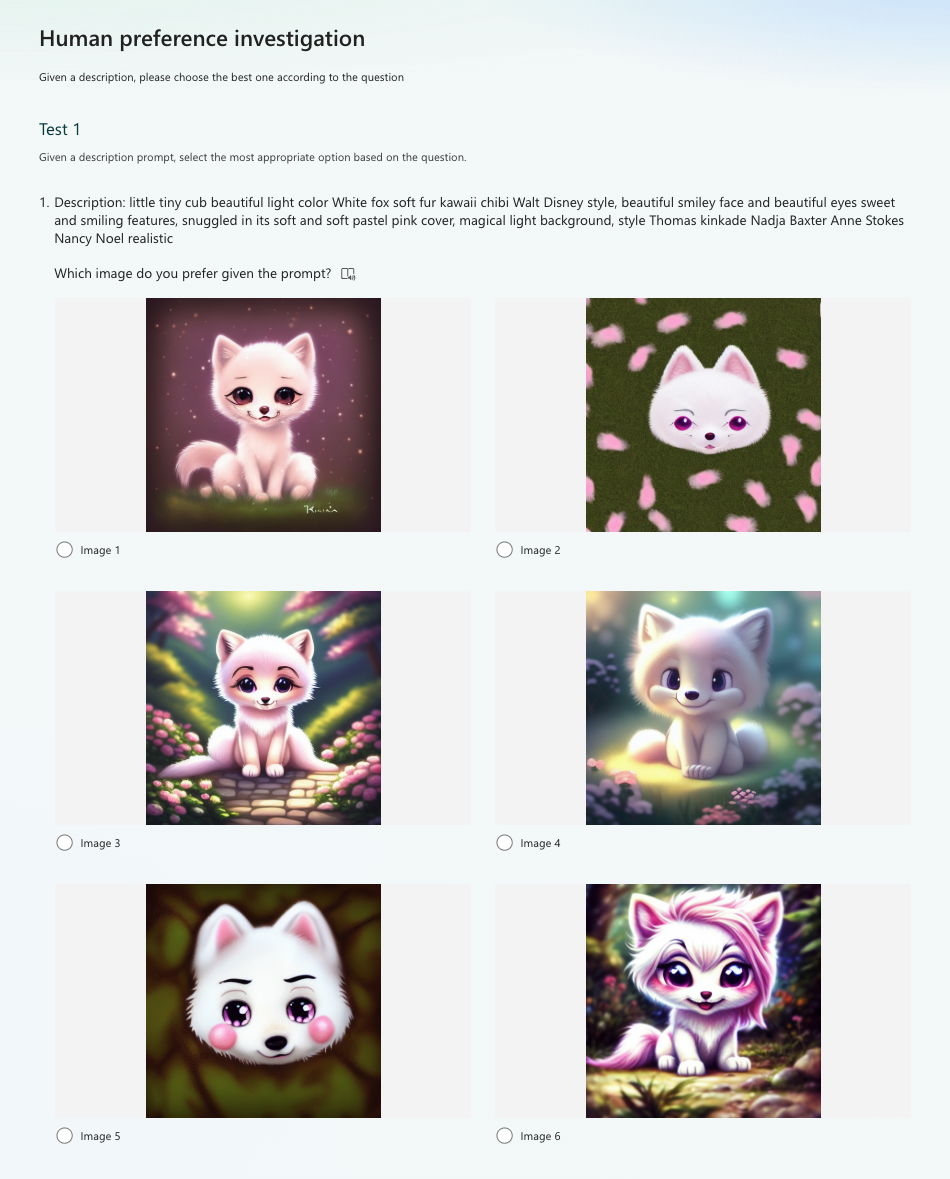}
\caption{The screenshot of human preference investigation: Which image do you prefer given the prompt?}
\label{fig:Q1}
\end{figure*}

\begin{figure*}[htbp]
\centering
 \includegraphics[width=0.9\linewidth]{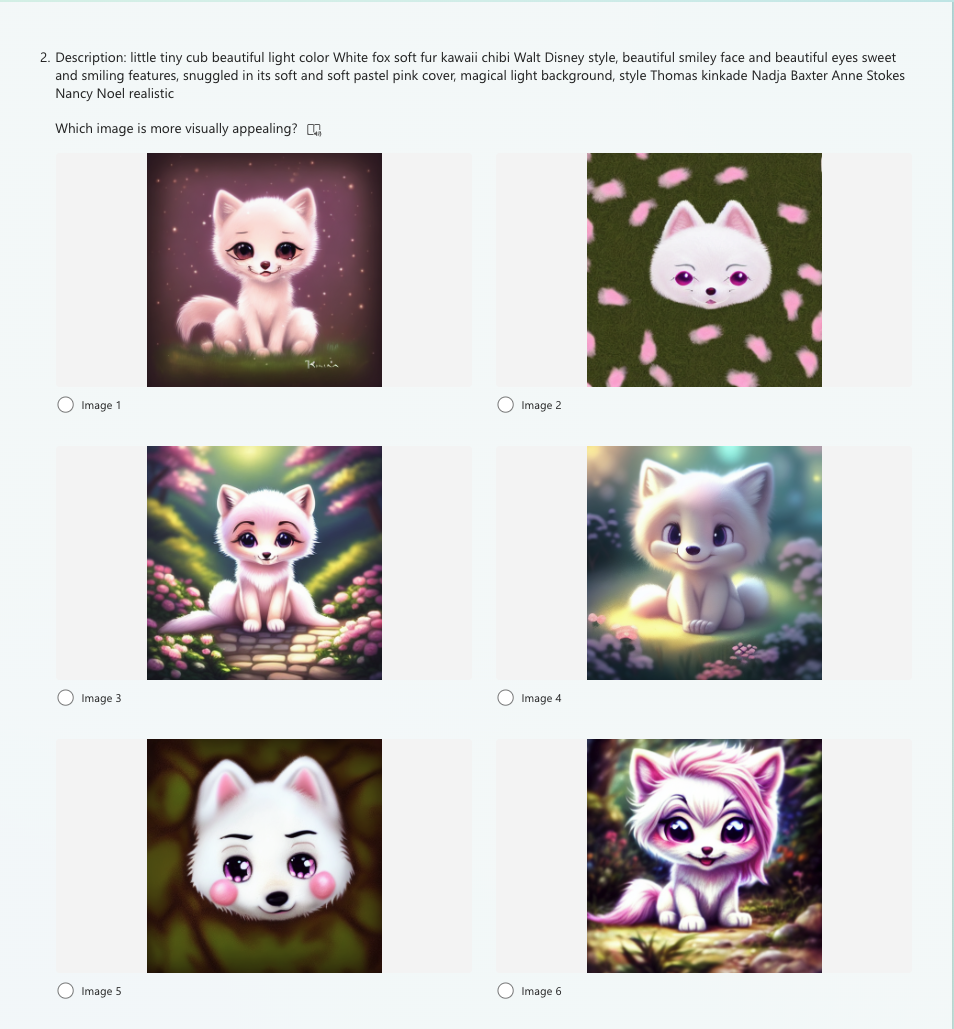}
\caption{The screenshot of human preference investigation: Which image is more visually appealing?}
\label{fig:Q2}
\end{figure*}

\begin{figure*}[htbp]
\centering
 \includegraphics[width=0.9\linewidth]{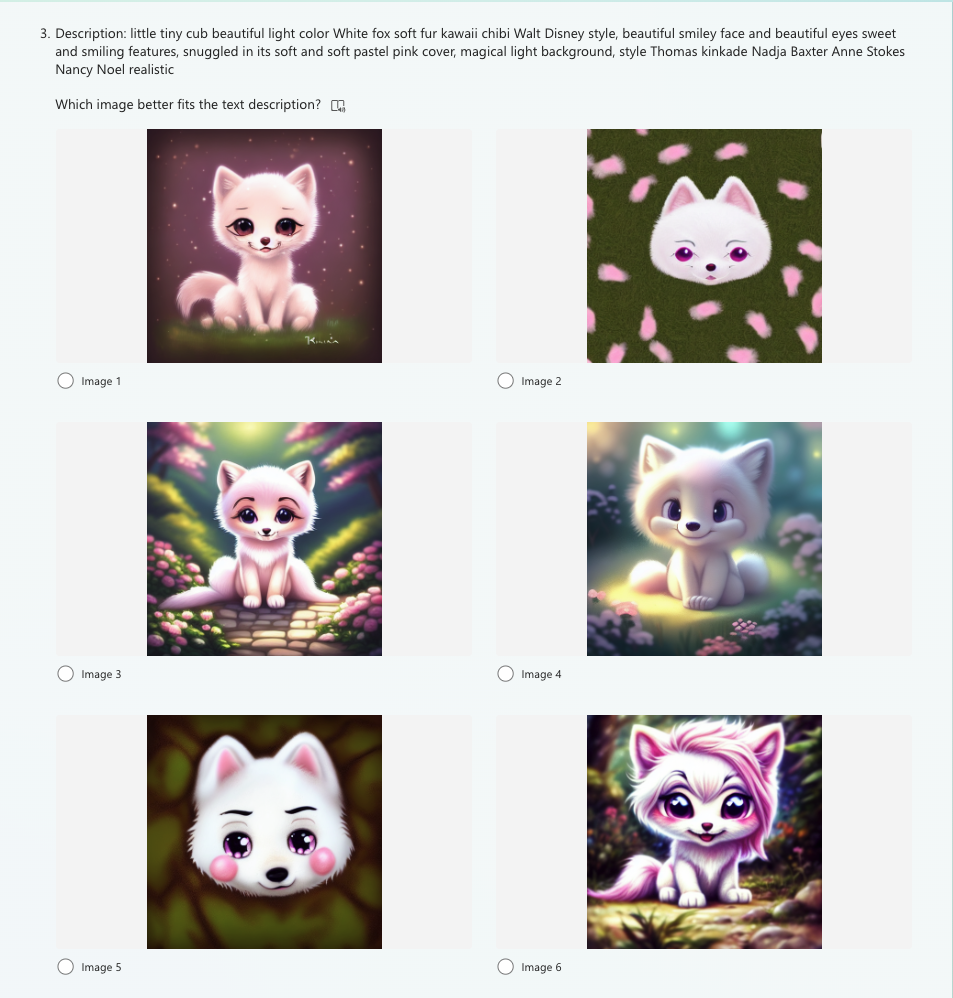}
\caption{The screenshot of human preference investigation: Which image better fits the text description?}
\label{fig:Q3}
\end{figure*}

We conduct the user study through survey forms, organizing the content into distinct sections based on each prompt. Each section is further divided into three partitions—Q1, Q2, and Q3—corresponding to \cref{fig:Q1}, \cref{fig:Q2}, \cref{fig:Q3}, respectively. In addition, investigators are recruited through an online platform, ensuring their anonymity. Each participant is required to have at least a bachelor’s degree and their privacy and identity are kept confidential throughout the entire process.

\begin{figure}[htp]
    \centering
    \begin{subfigure}[b]{0.22\textwidth}
        \includegraphics[width=\textwidth]{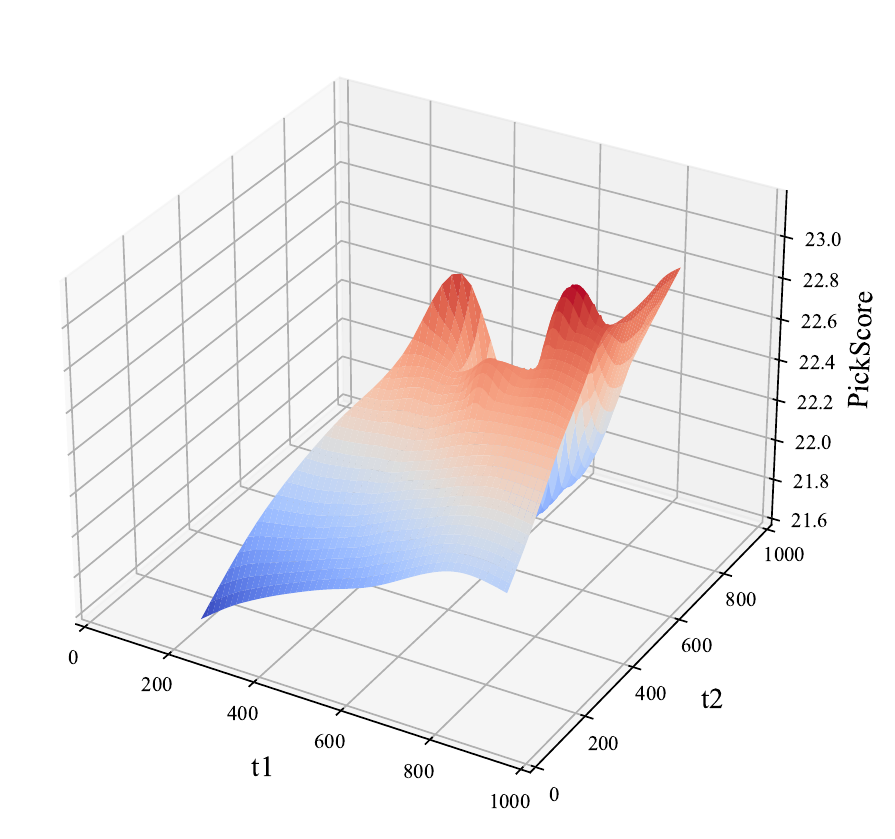}
        \caption{PickScore}
    \end{subfigure}
    \hspace{8pt} 
    \begin{subfigure}[b]{0.22\textwidth}
        \includegraphics[width=\textwidth]{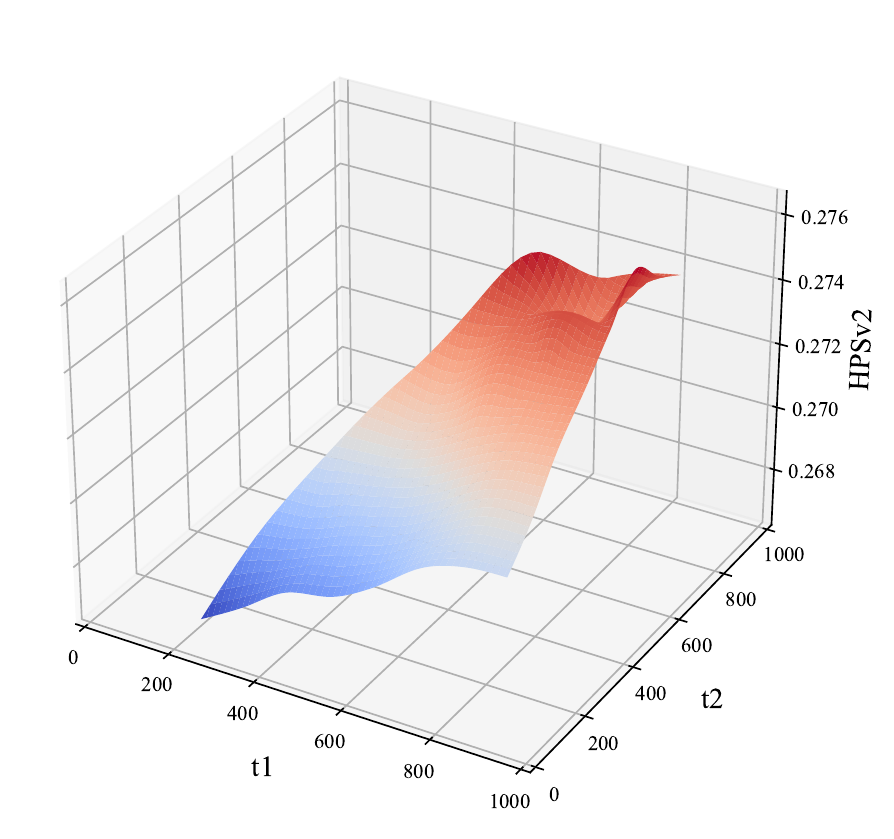}
        \caption{HPSv2}
    \end{subfigure}
    \vspace{8pt} 
    \begin{subfigure}[b]{0.22\textwidth}
        \includegraphics[width=\textwidth]{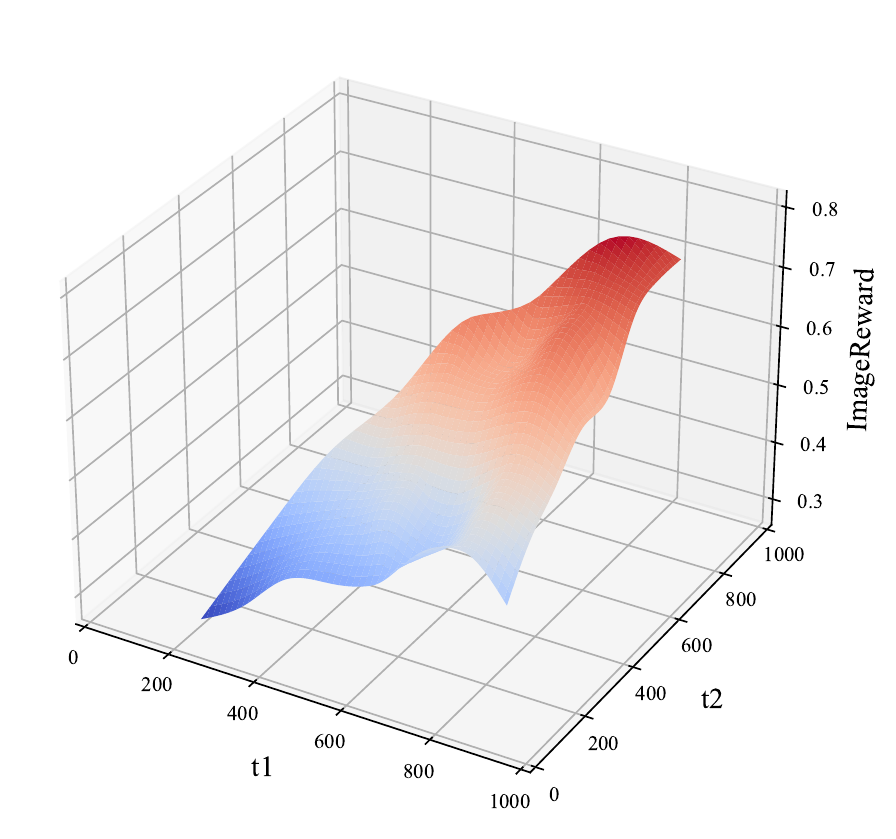}
        \caption{ImageReward}
    \end{subfigure}
    \hspace{8pt} 
    \begin{subfigure}[b]{0.22\textwidth}
        \includegraphics[width=\textwidth]{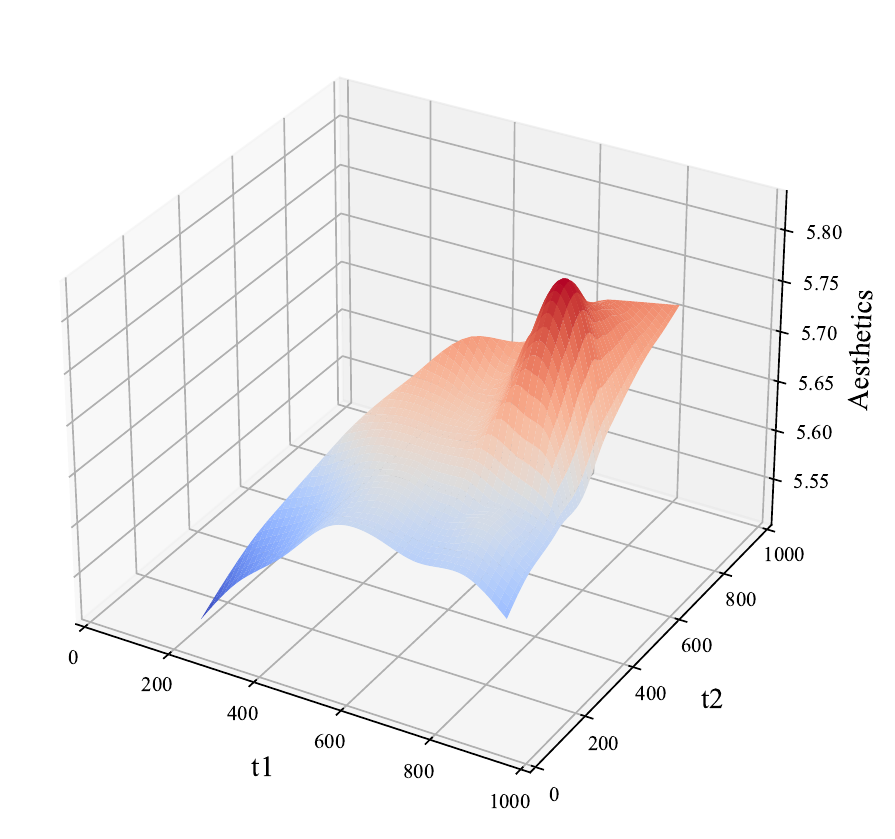}
        \caption{Aesthetics}
    \end{subfigure}
    \caption{Insensitivity to scheduling parameter $t_1$, $t_2$.}
    \label{fig:grid}
\end{figure}

\subsection{Additional Results on Dynamic Scheduling}
The scheduling weights (\cref{eq:dy_w}) are automatically adjusted based on content changes to balance the roles of $\mathcal{L}_A$ and $\mathcal{L}_R$ in different stages. In the early stage, $\mathcal{L}_A$ dominates due to content instability and noise (where $\mathcal{L}_R$'s guidance is weak). As semantics stabilize, $\mathcal{L}_R$ gains more weights, refining alignment with more detailed visual preferences. Thus, the method prioritizes semantic alignment first, then preference alignment. Based on the observed dynamics of $w_R$ and $w_A$ (trend and smoothness), we set the stage split $(t1, t2)$ as $(800, 500)$ for efficiency and simplicity. The model is insensitive to this hyperparameter, as confirmed by the grid-search analysis (\cref{fig:grid}).

\subsection{Additional Results on Dynamic Time Travel Steps}
In \cref{tab:SD15_dynamic_recurrent} in the main paper, we provide ablation study results of different time-travel steps. By comparison, our proposed dynamic recurrent strategy effectively achieves better performance on different metrics. We provide some visual comparison examples in \cref{fig:iter_ablation}, where we can observe that the visual qualities of the generated image are consistent with the numerical results. The proposed method with dynamic recurrent step scheduling is effective in producing better results with less time. With the proposed alignment objectives, the proposed method can also work well with less (and fixed) recurrent step numbers. In addition, compared with the other training-free methods with full-chain backpropagation methods, like DNO \cite{DNO} and PromptOpt \cite{PromptOpt}, require 370 and 280 seconds, respectively, our method can be more efficient and effective. Compared with the one-step approximation methods, such as FreeDom \cite{FreeDoM} (with 170 seconds), our method is also efficient and performs better on the results. While many training-free methods take more time to guide the denoising process, our method maintains strong performance in just 40s with fewer iterations
as demonstrated in \cref{tab:SD15_dynamic_recurrent} and \cref{fig:iter_ablation}. The whole generation process can be further accelerated by incorporating more efficient sampling processes, which is left as future work.

\section{Ethical and Social Impacts}
The development of DyMO, a training-free alignment framework for text-to-image diffusion models, brings ethical and social implications that require careful consideration to ensure responsible AI deployment. While our method enhances alignment with human preferences and promotes inclusivity, it also raises challenges such as mitigating biases, preserving privacy, and preventing misuse. DyMO relies on pre-trained models and publicly available datasets, which may encode societal biases or reinforce stereotypes. To address this, we emphasize the need for dataset diversity assessment, bias identification, and mechanisms to ensure inclusive and equitable representations. Privacy concerns are mitigated by advocating for anonymization of data and obtaining explicit consent for identifiable imagery. Additionally, we recognize the risks of misuse, such as generating harmful or misleading content, and propose safeguards like content moderation and ethical usage guidelines. Despite these challenges, DyMO holds the potential to advance social equality by improving accessibility and enabling personalized content generation for underrepresented groups. By balancing innovation with responsibility, we aim to democratize advanced generative techniques while upholding fairness, transparency, and inclusivity. Our commitment to responsible AI development underpins our efforts to address these concerns, ensuring that DyMO contributes positively to the field while minimizing potential risks.

\begin{table*}[t]
\centering
\small
\renewcommand{\arraystretch}{1.0}
\caption{Detailed prompts used for generated images in \cref{fig:first_image}.}
\label{tab:first_image_prompt_list}
\setlength{\tabcolsep}{5.0mm}{
\begin{tabular}{p{2cm}|p{12cm}}
\hline
\multicolumn{1}{c|}{Image} & \multicolumn{1}{c}{Prompt} \\ \hline

\multicolumn{1}{c|}{\scriptsize{\cref{fig:first_image}, Row 1, Col 1}} & 
\begin{minipage}[c]{\linewidth}
\vspace*{\fill}
\vspace*{0.5em}
\centering \scriptsize{Two monkeys are piloting an airplane.}
\vspace*{0.5em}
\vspace*{\fill}
\end{minipage} \\ \hline

\multicolumn{1}{c|}{\scriptsize{\cref{fig:first_image}, Row 1, Col 2}} & 
\begin{minipage}[c]{\linewidth}
\vspace*{\fill}
\vspace*{0.5em}
\centering \scriptsize{Award-winning Kawaii illustration of a cat samurai, holding two swords, background cyberpunk Styles, 4k, golden hour, cinematic light.}
\vspace*{0.5em}
\vspace*{\fill}
\end{minipage} \\ \hline

\multicolumn{1}{c|}{\scriptsize{\cref{fig:first_image}, Row 1, Col 3}} & 
\begin{minipage}[c]{\linewidth}
\vspace*{\fill}
\vspace*{0.5em}
\centering \scriptsize{crop top skinny russian 12 years old teen girl at the water mountain, HDR magazine photo.}
\vspace*{0.5em}
\vspace*{\fill}
\end{minipage} \\ \hline

\multicolumn{1}{c|}{\scriptsize{\cref{fig:first_image}, Row 1, Col 4}} & 
\begin{minipage}[c]{\linewidth}
\vspace*{\fill}
\vspace*{0.5em}
\centering \scriptsize{a tower of cheese.}
\vspace*{0.5em}
\vspace*{\fill}
\end{minipage} \\ \hline

\multicolumn{1}{c|}{\scriptsize{\cref{fig:first_image}, Row 1, Col 5}} & 
\begin{minipage}[c]{\linewidth}
\vspace*{\fill}
\vspace*{0.5em}
\centering \scriptsize{A painting of a koala wearing a princess dress and crown, with a confetti background.}
\vspace*{0.5em}
\vspace*{\fill}
\end{minipage} \\ \hline

\multicolumn{1}{c|}{\scriptsize{\cref{fig:first_image}, Row 1, Col 6}} & 
\begin{minipage}[c]{\linewidth}
\vspace*{\fill}
\vspace*{0.5em}
\centering \scriptsize{Disease Monitoring: Through big data technology, trends in specific diseases can be monitored and predicted, thus improving disease prevention and treatment effectiveness.}
\vspace*{0.5em}
\vspace*{\fill}
\end{minipage} \\ \hline

\multicolumn{1}{c|}{\scriptsize{\cref{fig:first_image}, Row 1, Col 7}} & 
\begin{minipage}[c]{\linewidth}
\vspace*{\fill}
\vspace*{0.5em}
\centering \scriptsize{Gnomes are playing music during Independence Day festivities in a forest near Lake George.}
\vspace*{0.5em}
\vspace*{\fill}
\end{minipage} \\ \hline

\multicolumn{1}{c|}{\scriptsize{\cref{fig:first_image}, Row 2, Col 1}} & 
\begin{minipage}[c]{\linewidth}
\vspace*{\fill}
\vspace*{0.5em}
\centering \scriptsize{paw patrol. 'This is some serious gourmet'. 2 dogs holding mugs.}
\vspace*{0.5em}
\vspace*{\fill}
\end{minipage} \\ \hline

\multicolumn{1}{c|}{\scriptsize{\cref{fig:first_image}, Row 2, Col 2}} & 
\begin{minipage}[c]{\linewidth}
\vspace*{\fill}
\vspace*{0.5em}
\centering \scriptsize{Harry potter as a cat, pixar style, octane render, HD, high-detail.}
\vspace*{0.5em}
\vspace*{\fill}
\end{minipage} \\ \hline

\multicolumn{1}{c|}{\scriptsize{\cref{fig:first_image}, Row 2, Col 3}} & 
\begin{minipage}[c]{\linewidth}
\vspace*{\fill}
\vspace*{0.5em}
\centering \scriptsize{A small green dinosaur toy with orange spots standing on its hind legs and roaring with its mouth open.}
\vspace*{0.5em}
\vspace*{\fill}
\end{minipage} \\ \hline

\multicolumn{1}{c|}{\scriptsize{\cref{fig:first_image}, Row 2, Col 3}} & 
\begin{minipage}[c]{\linewidth}
\vspace*{\fill}
\vspace*{0.5em}
\centering \scriptsize{Two cats watering roses in a greenhouse.}
\vspace*{0.5em}
\vspace*{\fill}
\end{minipage} \\ \hline

\multicolumn{1}{c|}{\scriptsize{\cref{fig:first_image}, Row 2, Col 4}} & 
\begin{minipage}[c]{\linewidth}
\vspace*{\fill}
\vspace*{0.5em}
\centering \scriptsize{Chic Fantasy Compositions, Ultra Detailed Artistic, Midnight Aura, Night Sky, Dreamy, Glowing, Glamour, Glimmer, Shadows, Oil On Canvas, Brush Strokes, Smooth, Ultra High Definition, 8k, Unreal Engine 5, Ultra Sharp Focus, Art By magali villeneuve, rossdraws, Intricate Artwork Masterpiece, Matte Painting Movie Poster.}
\vspace*{0.5em}
\vspace*{\fill}
\end{minipage} \\ \hline

\multicolumn{1}{c|}{\scriptsize{\cref{fig:first_image}, Row 2, Col 5}} & 
\begin{minipage}[c]{\linewidth}
\vspace*{\fill}
\vspace*{0.5em}
\centering \scriptsize{a toy poodle as a rocket scientist.}
\vspace*{0.5em}
\vspace*{\fill}
\end{minipage} \\ \hline

\multicolumn{1}{c|}{\scriptsize{\cref{fig:first_image}, Row 2, Col 5}} & 
\begin{minipage}[c]{\linewidth}
\vspace*{\fill}
\vspace*{0.5em}
\centering \scriptsize{A young woman witch cosplaying with a magic wand and broom, wearing boots, and posing in a full body shot with a detailed face.}
\vspace*{0.5em}
\vspace*{\fill}
\end{minipage} \\ \hline

\multicolumn{1}{c|}{\scriptsize{\cref{fig:first_image}, Row 3, Col 1}} & 
\begin{minipage}[c]{\linewidth}
\vspace*{\fill}
\vspace*{0.5em}
\centering \scriptsize{The image is a portrait of Homer Simpson as a Na'vi from Avatar, created with vibrant colors and highly detailed in a cinematic style reminiscent of romanticism by Eugene de Blaas and Ross Tran, available on Artstation with credits to Greg Rutkowski.}
\vspace*{0.5em}
\vspace*{\fill}
\end{minipage} \\ \hline

\multicolumn{1}{c|}{\scriptsize{\cref{fig:first_image}, Row 3, Col 2}} & 
\begin{minipage}[c]{\linewidth}
\vspace*{\fill}
\vspace*{0.5em}
\centering \scriptsize{Anthropomorphic beagle dog wearing steampunk time traveller outfit, clocks and large round window above, photoreal epic composition, old world deco, tv commercial, sebastian kruger, artem, epic lighting, by Heinz Anger, wow factor, aardman animations, blocking the sun, very artistic pose, alexander abdulov.}
\vspace*{0.5em}
\vspace*{\fill}
\end{minipage} \\ \hline

\multicolumn{1}{c|}{\scriptsize{\cref{fig:first_image}, Row 3, Col 3}} & 
\begin{minipage}[c]{\linewidth}
\vspace*{\fill}
\vspace*{0.5em}
\centering \scriptsize{A happy daffodil with big eyes, multiple leaf arms and vine legs, rendered in 3D Pixar style.}
\vspace*{0.5em}
\vspace*{\fill}
\end{minipage} \\ \hline

\multicolumn{1}{c|}{\scriptsize{\cref{fig:first_image}, Row 3, Col 4}} & 
\begin{minipage}[c]{\linewidth}
\vspace*{\fill}
\vspace*{0.5em}
\centering \scriptsize{A 3D Rendering of a cockatoo wearing sunglasses. The sunglasses have a deep black frame with bright pink lenses. Fashion photography, volumetric lighting, CG rendering.}
\vspace*{0.5em}
\vspace*{\fill}
\end{minipage} \\ \hline

\multicolumn{1}{c|}{\scriptsize{\cref{fig:first_image}, Row 3, Col 5}} & 
\begin{minipage}[c]{\linewidth}
\vspace*{\fill}
\vspace*{0.5em}
\centering \scriptsize{A slime monster.}
\vspace*{0.5em}
\vspace*{\fill}
\end{minipage} \\ \hline

\end{tabular}}
\end{table*}

\begin{table*}[htbp]
\centering
\small
\renewcommand{\arraystretch}{1.0}
\caption{Detailed prompts used for generated images in \cref{fig:sd15_compare}.}
\label{tab:sd15_compare_prompt_list}
\setlength{\tabcolsep}{5.0mm}{
\begin{tabular}{p{2cm}|p{12cm}}
\hline
\multicolumn{1}{c|}{Image} & \multicolumn{1}{c}{Prompt} \\ \hline

\multicolumn{1}{c|}{\scriptsize{\cref{fig:sd15_compare}, Row 1}} & 
\begin{minipage}[c]{\linewidth}
\vspace*{\fill}
\vspace*{0.5em}
\centering \scriptsize{A photo of a frog holding an apple while smiling in the forest.}
\vspace*{0.5em}
\vspace*{\fill}
\end{minipage} \\ \hline

\multicolumn{1}{c|}{\scriptsize{\cref{fig:sd15_compare}, Row 2}} & 
\begin{minipage}[c]{\linewidth}
\vspace*{\fill}
\vspace*{0.5em}
\centering \scriptsize{little tiny cub beautiful light color White fox soft fur kawaii chibi Walt Disney style, beautiful smiley face and beautiful eyes sweet and smiling features, snuggled in its soft and soft pastel pink cover, magical light background, style Thomas kinkade Nadja Baxter Anne Stokes Nancy Noel realistic.}
\vspace*{0.5em}
\vspace*{\fill}
\end{minipage} \\ \hline

\multicolumn{1}{c|}{\scriptsize{\cref{fig:sd15_compare}, Row 3}} & 
\begin{minipage}[c]{\linewidth}
\vspace*{\fill}
\vspace*{0.5em}
\centering \scriptsize{a gopro snapshot of an anthropomorphic cat dressed as a firefighter putting out a building fire.}
\vspace*{0.5em}
\vspace*{\fill}
\end{minipage} \\ \hline

\multicolumn{1}{c|}{\scriptsize{\cref{fig:sd15_compare}, Row 4}} & 
\begin{minipage}[c]{\linewidth}
\vspace*{\fill}
\vspace*{0.5em}
\centering \scriptsize{A swirling, multicolored portal emerges from the depths of an ocean of coffee, with waves of the rich liquid gently rippling outward. The portal engulfs a coffee cup, which serves as a gateway to a fantastical dimension. The surrounding digital art landscape reflects the colors of the portal, creating an alluring scene of endless possibilities.}
\vspace*{0.5em}
\vspace*{\fill}
\end{minipage} \\ \hline

\end{tabular}}
\end{table*}

\begin{table*}[t]
\centering
\small
\renewcommand{\arraystretch}{1.0}
\caption{Detailed prompts used for generated images in \cref{fig:sdxl_compare}.}
\label{tab:sdxl_compare_prompt_list}
\setlength{\tabcolsep}{5.0mm}{
\begin{tabular}{p{2cm}|p{12cm}}
\hline
\multicolumn{1}{c|}{Image} & \multicolumn{1}{c}{Prompt} \\ \hline

\multicolumn{1}{c|}{\scriptsize{\cref{fig:sdxl_compare}, Row 1}} & 
\begin{minipage}[c]{\linewidth}
\vspace*{\fill}
\vspace*{0.5em}
\centering \scriptsize{a golden retriever dressed like a General in the north army of the American Civil war. Portrait style, looking proud detailed 8k realistic super realistic Ultra HD cinematography photorealistic epic composition Unreal Engine Cinematic Color Grading portrait Photography UltraWide Angle Depth of Field hyperdetailed beautifully colorcoded insane details intricate details beautifully color graded Unreal Engine Editorial Photography Photography Photoshoot DOF Tilt Blur White Balance 32k SuperResolution Megapixel ProPhoto RGB VR Halfrear Lighting Backlight Natural Lighting Incandescent Optical Fiber Moody Lighting Cinematic Lighting Studio Lighting Soft Lighting Volumetric ContreJour Beautiful Lighting Accent Lighting Global Illumination Screen Space Global Illumination Ray Tracing Optics Scattering Glowing Shadows Rough Shimmering Ray Tracing Reflections Lumen Reflections Screen Space Reflections Diffraction Grading Chromatic Aberration GB Displacement Scan Lines Ray Traced Ray Tracing Ambient Occlusion AntiAliasing FKAA TXAA RTX SSAO Shaders.}
\vspace*{0.5em}
\vspace*{\fill}
\end{minipage} \\ \hline

\multicolumn{1}{c|}{\scriptsize{\cref{fig:sdxl_compare}, Row 2}} & 
\begin{minipage}[c]{\linewidth}
\vspace*{\fill}
\vspace*{0.5em}
\centering \scriptsize{Full body, a Super cute little girl, wearing cute little giraffe pajamas, Smile and look ahead, ultra detailed sky blue eyes, 8k bright front lighting, fine luster, ultra detail, hyper detailed 3D rendering s750.}
\vspace*{0.5em}
\vspace*{\fill}
\end{minipage} \\ \hline

\multicolumn{1}{c|}{\scriptsize{\cref{fig:sdxl_compare}, Row 3}} & 
\begin{minipage}[c]{\linewidth}
\vspace*{\fill}
\vspace*{0.5em}
\centering \scriptsize{A smiling beautiful sorceress wearing a high necked blue suit surrounded by swirling rainbow aurora, hyper-realistic, cinematic, post-production.}
\vspace*{0.5em}
\vspace*{\fill}
\end{minipage} \\ \hline

\multicolumn{1}{c|}{\scriptsize{\cref{fig:sdxl_compare}, Row 4}} & 
\begin{minipage}[c]{\linewidth}
\vspace*{\fill}
\vspace*{0.5em}
\centering \scriptsize{a white polar bear cub wearing sunglasses sits in a meadow with flowers.}
\vspace*{0.5em}
\vspace*{\fill}
\end{minipage} \\ \hline

\end{tabular}}
\end{table*}

\begin{table*}[t]
\centering
\small
\renewcommand{\arraystretch}{1.0}
\caption{Detailed prompts used for generated images in \cref{fig:compare_sd15_more_our}.}
\label{tab:sd15_compare_more_prompt_list}
\setlength{\tabcolsep}{5.0mm}{
\begin{tabular}{p{2cm}|p{12cm}}
\hline
\multicolumn{1}{c|}{Image} & \multicolumn{1}{c}{Prompt} \\ \hline
\multicolumn{1}{c|}{\scriptsize{\cref{fig:compare_sd15_more_our}, Row 1}} & 
\begin{minipage}[c]{\linewidth}
\vspace*{\fill}
\vspace*{0.5em}
\centering \scriptsize{On the Mid-Autumn Festival, the bright full moon hangs in the night sky. A quaint pavilion is illuminated by dim lights, resembling a beautiful scenery in a painting. Camera type: close-up. Camera lens type: telephoto. Time of day: night. Style of lighting: bright. Film type: ancient style. HD. }
\vspace*{0.5em}
\vspace*{\fill}
\end{minipage} \\ \hline

\multicolumn{1}{c|}{\scriptsize{\cref{fig:compare_sd15_more_our}, Row 2}} & 
\begin{minipage}[c]{\linewidth}
\vspace*{\fill}
\vspace*{0.5em}
\centering \scriptsize{A 3D Rendering of a cockatoo wearing sunglasses. The sunglasses have a deep black frame with bright pink lenses. Fashion photography, volumetric lighting, CG rendering.}
\vspace*{0.5em}
\vspace*{\fill}
\end{minipage} \\ \hline

\multicolumn{1}{c|}{\scriptsize{\cref{fig:compare_sd15_more_our}, Row 3}} & 
\begin{minipage}[c]{\linewidth}
\vspace*{\fill}
\vspace*{0.5em}
\centering \scriptsize{A painting of a koala wearing a princess dress and crown, with a confetti background.}
\vspace*{0.5em}
\vspace*{\fill}
\end{minipage} \\ \hline

\multicolumn{1}{c|}{\scriptsize{\cref{fig:compare_sd15_more_our}, Row 4}} & 
\begin{minipage}[c]{\linewidth}
\vspace*{\fill}
\vspace*{0.5em}
\centering \scriptsize{A rock formation in the shape of a horse, insanely detailed.}
\vspace*{0.5em}
\vspace*{\fill}
\end{minipage} \\ \hline

\multicolumn{1}{c|}{\scriptsize{\cref{fig:compare_sd15_more_our}, Row 5}} & 
\begin{minipage}[c]{\linewidth}
\vspace*{\fill}
\vspace*{0.5em}
\centering \scriptsize{New York city skyline during the day with a huge red moon in the sky.}
\vspace*{0.5em}
\vspace*{\fill}
\end{minipage} \\ \hline

\multicolumn{1}{c|}{\scriptsize{\cref{fig:compare_sd15_more_our}, Row 6}} & 
\begin{minipage}[c]{\linewidth}
\vspace*{\fill}
\vspace*{0.5em}
\centering \scriptsize{A stained glass vase with flowers in front of a window.}
\vspace*{0.5em}
\vspace*{\fill}
\end{minipage} \\ \hline

\multicolumn{1}{c|}{\scriptsize{\cref{fig:compare_sd15_more_our}, Row 7}} & 
\begin{minipage}[c]{\linewidth}
\vspace*{\fill}
\vspace*{0.5em}
\centering \scriptsize{A white polar bear cub wearing sunglasses sits in a meadow with flowers.}
\vspace*{0.5em}
\vspace*{\fill}
\end{minipage} \\ \hline

\multicolumn{1}{c|}{\scriptsize{\cref{fig:compare_sd15_more_our}, Row 8}} & 
\begin{minipage}[c]{\linewidth}
\vspace*{\fill}
\vspace*{0.5em}
\centering \scriptsize{Warsaw cyberpunk style at night.}
\vspace*{0.5em}
\vspace*{\fill}
\end{minipage} \\ \hline

\multicolumn{1}{c|}{\scriptsize{\cref{fig:compare_sd15_more_our}, Row 9}} & 
\begin{minipage}[c]{\linewidth}
\vspace*{\fill}
\vspace*{0.5em}
\centering \scriptsize{astronaut in space with a two handed sword in plate armor in front of the earth.}
\vspace*{0.5em}
\vspace*{\fill}
\end{minipage} \\ \hline

\multicolumn{1}{c|}{\scriptsize{\cref{fig:compare_sd15_more_our}, Row 10}} & 
\begin{minipage}[c]{\linewidth}
\vspace*{\fill}
\vspace*{0.5em}
\centering \scriptsize{a slice of pizza floating through space with stars in the background.}
\vspace*{0.5em}
\vspace*{\fill}
\end{minipage} \\ \hline

\multicolumn{1}{c|}{\scriptsize{\cref{fig:compare_sd15_more_our}, Row 11}} & 
\begin{minipage}[c]{\linewidth}
\vspace*{\fill}
\vspace*{0.5em}
\centering \scriptsize{a cute bunny wear detailed metal armour.}
\vspace*{0.5em}
\vspace*{\fill}
\end{minipage} \\ \hline

\end{tabular}}
\end{table*}

\begin{table*}[t]
\centering
\small
\renewcommand{\arraystretch}{1.0}
\caption{Detailed prompts used for generated images in \cref{fig:compare_sdxl_more1} and \cref{fig:compare_sdxl_more2}.}
\label{tab:sdxl_compare_more_prompt_list}
\setlength{\tabcolsep}{5.0mm}{
\begin{tabular}{p{2cm}|p{12cm}}
\hline
\multicolumn{1}{c|}{Image} & \multicolumn{1}{c}{Prompt} \\ \hline
\multicolumn{1}{c|}{\scriptsize{\cref{fig:compare_sdxl_more1}, Row 1}} & 
\begin{minipage}[c]{\linewidth}
\vspace*{\fill}
\vspace*{0.5em}
\centering \scriptsize{A panda bear as a mad scientist.}
\vspace*{0.5em}
\vspace*{\fill}
\end{minipage} \\ \hline

\multicolumn{1}{c|}{\scriptsize{\cref{fig:compare_sdxl_more1}, Row 2}} & 
\begin{minipage}[c]{\linewidth}
\vspace*{\fill}
\vspace*{0.5em}
\centering \scriptsize{Green laser in space among galaxies forest.}
\vspace*{0.5em}
\vspace*{\fill}
\end{minipage} \\ \hline

\multicolumn{1}{c|}{\scriptsize{\cref{fig:compare_sdxl_more1}, Row 3}} & 
\begin{minipage}[c]{\linewidth}
\vspace*{\fill}
\vspace*{0.5em}
\centering \scriptsize{a slice of pizza floating through space with stars in the background.}
\vspace*{0.5em}
\vspace*{\fill}
\end{minipage} \\ \hline

\multicolumn{1}{c|}{\scriptsize{\cref{fig:compare_sdxl_more1}, Row 4}} & 
\begin{minipage}[c]{\linewidth}
\vspace*{\fill}
\vspace*{0.5em}
\centering \scriptsize{disney league of legends splash art of space marine donald trump.}
\vspace*{0.5em}
\vspace*{\fill}
\end{minipage} \\ \hline

\multicolumn{1}{c|}{\scriptsize{\cref{fig:compare_sdxl_more1}, Row 5}} & 
\begin{minipage}[c]{\linewidth}
\vspace*{\fill}
\vspace*{0.5em}
\centering \scriptsize{one piece anime cover.}
\vspace*{0.5em}
\vspace*{\fill}
\end{minipage} \\ \hline

\multicolumn{1}{c|}{\scriptsize{\cref{fig:compare_sdxl_more1}, Row 6}} & 
\begin{minipage}[c]{\linewidth}
\vspace*{\fill}
\vspace*{0.5em}
\centering \scriptsize{cyberpunk wizard.}
\vspace*{0.5em}
\vspace*{\fill}
\end{minipage} \\ \hline

\multicolumn{1}{c|}{\scriptsize{\cref{fig:compare_sdxl_more1}, Row 7}} & 
\begin{minipage}[c]{\linewidth}
\vspace*{\fill}
\vspace*{0.5em}
\centering \scriptsize{The gate to the eternal kingdom of angels, fantasy, digital painting, HD, detailed.}
\vspace*{0.5em}
\vspace*{\fill}
\end{minipage} \\ \hline

\multicolumn{1}{c|}{\scriptsize{\cref{fig:compare_sdxl_more1}, Row 8}} & 
\begin{minipage}[c]{\linewidth}
\vspace*{\fill}
\vspace*{0.5em}
\centering \scriptsize{Hirundo a data startup logo.}
\vspace*{0.5em}
\vspace*{\fill}
\end{minipage} \\ \hline

\multicolumn{1}{c|}{\scriptsize{\cref{fig:compare_sdxl_more1}, Row 9}} & 
\begin{minipage}[c]{\linewidth}
\vspace*{\fill}
\vspace*{0.5em}
\centering \scriptsize{a dragon vinyl toy in a fighting pose.}
\vspace*{0.5em}
\vspace*{\fill}
\end{minipage} \\ \hline

\multicolumn{1}{c|}{\scriptsize{\cref{fig:compare_sdxl_more1}, Row 10}} & 
\begin{minipage}[c]{\linewidth}
\vspace*{\fill}
\vspace*{0.5em}
\centering \scriptsize{Psychedelic synesthesia complex 3d rendered ethereal 8 complex shapes in different sizes fractal futuristic geometry POV replication of future riddim.}
\vspace*{0.5em}
\vspace*{\fill}
\end{minipage} \\ \hline

\multicolumn{1}{c|}{\scriptsize{\cref{fig:compare_sdxl_more1}, Row 11}} & 
\begin{minipage}[c]{\linewidth}
\vspace*{\fill}
\vspace*{0.5em}
\centering \scriptsize{masterpiece portrait of Kristen Stewart standing in the movie set, film photography, dark atmosphere, sharp focus, photographed by Annie Leibovitz.}
\vspace*{0.5em}
\vspace*{\fill}
\end{minipage} \\ \hline

\multicolumn{1}{c|}{\scriptsize{\cref{fig:compare_sdxl_more1}, Row 12}} & 
\begin{minipage}[c]{\linewidth}
\vspace*{\fill}
\vspace*{0.5em}
\centering \scriptsize{mechanical bee flying in nature, electronics, motors, wires, buttons, lcd, led instead of eyes, antennas instead of feet.}
\vspace*{0.5em}
\vspace*{\fill}
\end{minipage} \\ \hline

\multicolumn{1}{c|}{\scriptsize{\cref{fig:compare_sdxl_more2}, Row 1}} & 
\begin{minipage}[c]{\linewidth}
\vspace*{\fill}
\vspace*{0.5em}
\centering \scriptsize{At Song dynasty, a pretty woman in chinese was walking along the river.}
\vspace*{0.5em}
\vspace*{\fill}
\end{minipage} \\ \hline

\multicolumn{1}{c|}{\scriptsize{\cref{fig:compare_sdxl_more2}, Row 2}} & 
\begin{minipage}[c]{\linewidth}
\vspace*{\fill}
\vspace*{0.5em}
\centering \scriptsize{spiral mountains, stars, clouds within spheres, green pine trees, chill, calmness, peace, eternity, beauty, ernst haeckel, maria sibylla merian, tristan eaton, victo ngai, artgerm, rhads, ross draws, kaethe butcher, hajime sorayama, greg tocchini, virgil finlay, subtle vignette, volumetric lights, pixiv, by ilya kuvshinov, octane render, 4k, 8k.}
\vspace*{0.5em}
\vspace*{\fill}
\end{minipage} \\ \hline

\multicolumn{1}{c|}{\scriptsize{\cref{fig:compare_sdxl_more2}, Row 3}} & 
\begin{minipage}[c]{\linewidth}
\vspace*{\fill}
\vspace*{0.5em}
\centering \scriptsize{Dot matrix, pointillism, seurat, signac, frazetta, brom, Zdzisław Beksiński, Moebius, Egon Schiele, art nouveau, a being made out of pure golden light. angelic and graceful. gold foil inlay with erratic shapes and geometric patterns. black and gold and white and orange. in the style of alphonse mucha and amanda guse and android jones. tarot card style symmetry.}
\vspace*{0.5em}
\vspace*{\fill}
\end{minipage} \\ \hline

\multicolumn{1}{c|}{\scriptsize{\cref{fig:compare_sdxl_more2}, Row 4}} & 
\begin{minipage}[c]{\linewidth}
\vspace*{\fill}
\vspace*{0.5em}
\centering \scriptsize{A dark hole in the sky , pont of view from a beach, masterpiece.}
\vspace*{0.5em}
\vspace*{\fill}
\end{minipage} \\ \hline

\multicolumn{1}{c|}{\scriptsize{\cref{fig:compare_sdxl_more2}, Row 5}} & 
\begin{minipage}[c]{\linewidth}
\vspace*{\fill}
\vspace*{0.5em}
\centering \scriptsize{Cute and adorable cartoon rabbit baby rhea facing the camera, fantasy, dreamlike, surrealism, super cute, trending on artstationm volumetric light, cinematic, post processing, 8K.}
\vspace*{0.5em}
\vspace*{\fill}
\end{minipage} \\ \hline

\multicolumn{1}{c|}{\scriptsize{\cref{fig:compare_sdxl_more2}, Row 6}} & 
\begin{minipage}[c]{\linewidth}
\vspace*{\fill}
\vspace*{0.5em}
\centering \scriptsize{magic caster.}
\vspace*{0.5em}
\vspace*{\fill}
\end{minipage} \\ \hline

\multicolumn{1}{c|}{\scriptsize{\cref{fig:compare_sdxl_more2}, Row 7}} & 
\begin{minipage}[c]{\linewidth}
\vspace*{\fill}
\vspace*{0.5em}
\centering \scriptsize{Cadillac El Dorado de style Badass Steampunk dans une vieille rue pavée, trending on artstation, sharp focus, studio photo, intricate details, highly detailed, by greg rutkowski.}
\vspace*{0.5em}
\vspace*{\fill}
\end{minipage} \\ \hline

\multicolumn{1}{c|}{\scriptsize{\cref{fig:compare_sdxl_more2}, Row 8}} & 
\begin{minipage}[c]{\linewidth}
\vspace*{\fill}
\vspace*{0.5em}
\centering \scriptsize{character sheet,The little girl riding an electric scooter bike, in a beautiful anime scene by Hayao Miyazaki: a snowy Tokyo city with massive Miyazaki clouds floating in the blue sky, enchanting snowscapes of the city with bright sunlight, Miyazaki's landscape imagery, Japanese art, 16:9.}
\vspace*{0.5em}
\vspace*{\fill}
\end{minipage} \\ \hline

\multicolumn{1}{c|}{\scriptsize{\cref{fig:compare_sdxl_more2}, Row 9}} & 
\begin{minipage}[c]{\linewidth}
\vspace*{\fill}
\vspace*{0.5em}
\centering \scriptsize{3D digital illustration, Burger with wheels speeding on the race track, supercharged, detailed, hyperrealistic, 4K.}
\vspace*{0.5em}
\vspace*{\fill}
\end{minipage} \\ \hline

\multicolumn{1}{c|}{\scriptsize{\cref{fig:compare_sdxl_more2}, Row 10}} & 
\begin{minipage}[c]{\linewidth}
\vspace*{\fill}
\vspace*{0.5em}
\centering \scriptsize{magic the gathering, anthro furry knight adventurer, showcase promo full art. Painted impressionist style.}
\vspace*{0.5em}
\vspace*{\fill}
\end{minipage} \\ \hline

\multicolumn{1}{c|}{\scriptsize{\cref{fig:compare_sdxl_more2}, Row 11}} & 
\begin{minipage}[c]{\linewidth}
\vspace*{\fill}
\vspace*{0.5em}
\centering \scriptsize{hundreds of sith warships in space facing viewer, symmetrical, centered, front view, highly detailed, centered, digital painting, ultradetailed, artstation, digital painting, cgsociety, octane render, sharp focus, illustration, cinematic lighting, 8k hd hyper realistic, intricate, lifelike, golden hour, highly detailed, art by ralph mcquarrie, James Ferdinand Knab, William O'Keefe, Boris Vallejo, Peter Kemp, Joshy Lee, Otto Schmit, and Aja RICKO.}
\vspace*{0.5em}
\vspace*{\fill}
\end{minipage} \\ \hline

\multicolumn{1}{c|}{\scriptsize{\cref{fig:compare_sdxl_more2}, Row 12}} & 
\begin{minipage}[c]{\linewidth}
\vspace*{\fill}
\vspace*{0.5em}
\centering \scriptsize{Lion dragon hybrid.}
\vspace*{0.5em}
\vspace*{\fill}
\end{minipage} \\ \hline

\end{tabular}}
\end{table*}

\begin{figure*}[htbp]
\centering
\begin{tcolorbox}[colback=gray!20, colframe=gray!80, boxrule=0.5pt, rounded corners, width=0.9\textwidth]
\linespread{1.1}\selectfont

\texttt{\{} \\
\texttt{"prompt": "a gopro snapshot of an anthropomorphic cat dressed as a firefighter putting out a building fire", } \\
\texttt{"Graph": [} \\
\texttt{\{"cat": ["anthropomorphic"]\},} \\
\texttt{\{"firefighter": []\}, } \\
\texttt{\{"fire": ["building"]\} ]} \\
\texttt{\}}

\vspace{3em}

\texttt{\{} \\
\texttt{"prompt": "a white polar bear cub wearing sunglasses sits in a meadow with flowers.", } \\
\texttt{"Graph": [} \\
\texttt{\{"bear": ["white", "cub"]\},} \\
\texttt{\{"sunglasses": []\}, } \\
\texttt{\{"meadow": []\}, } \\
\texttt{\{"flowers": []\} ]} \\
\texttt{\}}

\vspace{3em}

\texttt{\{} \\
\texttt{"prompt": "a photo of a frog holding an apple while smiling in the forest", } \\
\texttt{"Graph": [} \\
\texttt{\{"frog": ["smiling"]\}, } \\
\texttt{\{"apple": []\}, } \\
\texttt{\{"forest": []\} ]} \\
\texttt{\}}

\vspace{3em}

\texttt{\{} \\
\texttt{"prompt": "A swirling, multicolored portal emerges from the depths of an ocean of coffee, with waves of the rich liquid gently rippling outward. The portal engulfs a coffee cup, which serves as a gateway to a fantastical dimension. The surrounding digital art landscape reflects the colors of the portal, creating an alluring scene of endless possibilities.", } \\
\texttt{"Graph": [} \\
\texttt{\{"portal": ["swirling", "multicolored", "emerges"]\}, } \\
\texttt{\{"ocean": ["depths", "coffee", "rich"]\}, } \\
\texttt{\{"waves": ["rippling", "gently", "outward"]\}, } \\
\texttt{\{"cup": ["coffee", "gateway"]\} ]} \\
\texttt{\}}

\end{tcolorbox}

\caption{Some examples of text semantic graph.}
\label{fig:graph_example}
\end{figure*}


\end{document}